\pdfoutput=1
\documentclass[journal]{IEEEtran}

\usepackage{amsmath, amssymb, amsthm}
\usepackage{bm}

\usepackage{graphicx}
\usepackage[caption=false,font=footnotesize]{subfig}
\usepackage{array}
\usepackage{booktabs}
\usepackage{multirow}
\usepackage{rotating}

\usepackage{tikz}
\usetikzlibrary{positioning, arrows.meta, shapes, decorations.pathreplacing, calc, fit, backgrounds}

\usepackage{cite}

\usepackage{xcolor}
\usepackage{colortbl}
\definecolor{ourrow}{gray}{0.94}
\definecolor{refrow}{gray}{0.80}
\definecolor{priorrow}{rgb}{0.93,0.93,0.74}
\definecolor{genericrow}{rgb}{0.80,0.80,0.55}
\usepackage{url}
\usepackage{placeins}
\usepackage{float}

\usepackage[hidelinks]{hyperref}
\usepackage[capitalize]{cleveref}
\crefname{figure}{Fig.}{Figs.}
\Crefname{figure}{Fig.}{Figs.}

\makeatletter
\def\input@path{{./}{./fig/}{./tab/}}
\makeatother
\graphicspath{{fig/}{fig/img/}}

\newcommand{\norm}[1]{\left\lVert#1\right\rVert}

\newcommand{\Ltwo}{\ell_{2}}

\theoremstyle{definition}

\theoremstyle{remark}

\title{\textit{Damnatio Memoriae}: Adversarially and Selectively Forgetting Identities in the Embedding Space of Face Recognition Models}

\author{\"Unsal \"Ozt\"urk,~Vedrana Krivoku{\'c}a Hahn,~Sushil Bhattacharjee,~and~S\'ebastien~Marcel%
\thanks{\"U. \"Ozt\"urk, V. Krivoku{\'c}a Hahn, S. Bhattacharjee and S. Marcel are with the Idiap Research Institute, Martigny, Switzerland. S. Marcel is also with UNIL, Lausanne, Switzerland.}%
\thanks{Corresponding author: \"U. \"Ozt\"urk (e-mail: unsal.ozturk@idiap.ch).}%
\thanks{This work might be submitted to the IEEE for possible publication. Copyright may be transferred without notice, after which this version may no longer be accessible.}%
}

\begin{document}

\bstctlcite{IEEEexample:BSTcontrol}

\maketitle

\begin{abstract}
\noindent
A face recognition model links two images of a person recorded on separate occasions when their embedding similarity exceeds an operating threshold. We consider making chosen identities unlinkable across separate occasions while the model remains in service for the rest of the population. Deleting their images and retraining does not achieve this, since the model recognises identities never observed in training. Therefore, the embedding space must be altered against these identities, the process of which we call open-set adversarial forgetting. We propose three loss functions, one that disperses an identity's embeddings from their centroid, and two that map each image onto its own near-orthogonal target, learnt with the classifier head or fixed in advance as an almost-orthonormal frame. Each is fine-tuned alongside the classification objective on a subset of each identity's images. We evaluate them against four methods from prior work in verification and identification, at two forget scales and three backbones. Every loss acting on the embedding geometry makes the forget identities nearly unidentifiable. The orthonormal frame alone achieves strong forgetting, which holds wherever an image of that subset enters the comparison and leaves distinct forget identities unlinkable. It also surpasses a concurrent unsupervised method at a higher retain rate.
\end{abstract}


\IEEEpeerreviewmaketitle

\section{Introduction}
\label{sec:intro}

\begin{figure}[t]
  \centering
  \includegraphics[width=0.78\columnwidth]{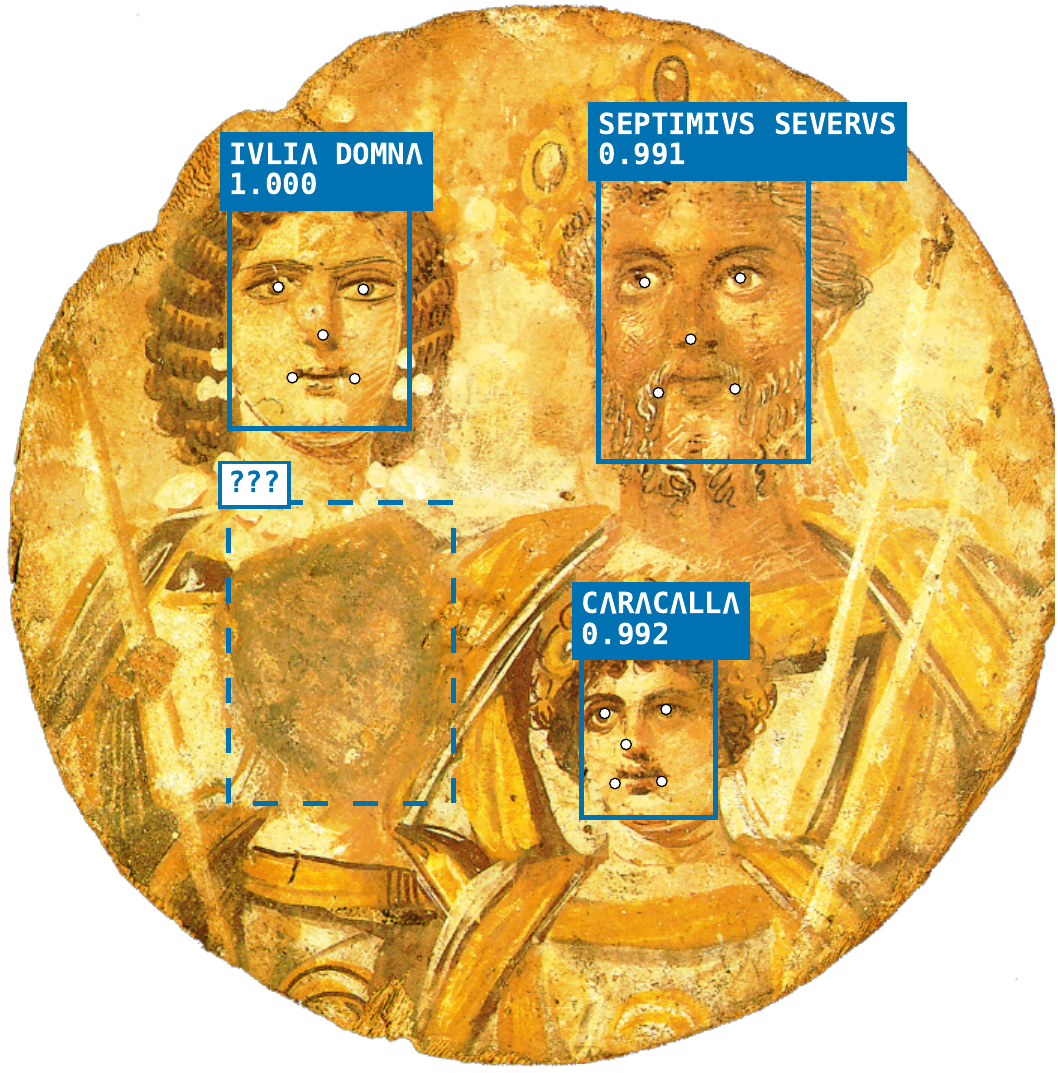}
  \caption{An example of damnatio memoriae, the Roman practice of erasing an individual from public memory. In the Severan Tondo~\cite{severan_tondo} the face of Geta has been scraped from the panel. We pursue the same effect in the \textbf{embedding space of the face recognition model}, such that selected identities are ``scraped'' in an open-set manner.}
  \label{fig:teaser}
\end{figure}

Modern face recognition backbones are trained with angular-margin classification losses such as SphereFace~\cite{DBLP:conf/cvpr/LiuWYLRS17}, CosFace~\cite{DBLP:conf/cvpr/WangWZJGZL018}, ArcFace~\cite{DBLP:conf/cvpr/DengGXZ19}, and AdaFace~\cite{DBLP:conf/cvpr/Kim0L22} on web-scale datasets containing hundreds of thousands of identities~\cite{DBLP:conf/cvpr/ZhuHDY0CZYLD021,DBLP:conf/fgr/CaoSXPZ18}, a family that consolidates earlier embedding-space objectives such as the triplet loss of FaceNet~\cite{DBLP:conf/cvpr/SchroffKP15} and the centre loss of Wen et al.~\cite{DBLP:conf/eccv/WenZL016}.
These models learn an $\Ltwo$-normalised embedding space which organises faces by identity through angular distances, such that the cosine similarity of two mated face embeddings is higher than that of non-mated embeddings, and the two comparison types therefore follow distinguishable similarity distributions. Whether two face images belong to the same identity can therefore be determined by comparing their cosine similarity to a calibrated operating threshold separating the two distributions. Once such a model is deployed, data-protection law may require a particular individual to be forgotten on request, and what it means for the model to have done so has more than one reasonable answer.

The unlearning literature uses ``unlearning'' in two structurally distinct senses, which we disambiguate as follows. The first and classical sense~\cite{DBLP:conf/sp/CaoY15,DBLP:conf/sp/BourtouleCCJTZL21,DBLP:conf/nips/GinartGVZ19,DBLP:conf/icml/GuoGHM20,DBLP:conf/nips/SekhariAKS21,DBLP:conf/ndss/WarneckePWR23,DBLP:conf/aaai/FosterSB24}, in which the majority of existing methods operate, considers a model to be ``unlearned'' if its post-unlearning weight distribution is statistically indistinguishable from that of a model retrained from scratch on the same training data with the ``forget'' set removed. This property is assessed against the retrained reference through membership-inference attacks~\cite{DBLP:journals/corr/abs-2403-01218} or representation-similarity probes~\cite{DBLP:journals/corr/abs-2503-06991,DBLP:journals/corr/abs-2504-14798}. The second sense considers a model to be unlearned if its verification behaviour on the chosen identities is degraded relative to that of the original model, while its behaviour on every other identity is preserved.

In the context of open-set face recognition, the first sense of unlearning does not necessarily imply that the chosen identities become unverifiable. The face embedding manifold is shaped jointly by all its training identities, and generalises to identities never observed at training time, given sufficient training data. It follows that a model retrained from scratch with samples of a particular identity removed could still be used to verify pairs of face images of that individual. We adopt the second sense as the operational target of this work and henceforth refer to it as \textit{adversarial forgetting}, in which the embedding space is altered against the chosen identities so that their images are no longer verified against one another, while the rest of the population is left intact.

Our contributions are:
\begin{enumerate}
     \item We formulate adversarial forgetting as degrading the recognition performance of the model deliberately on the chosen identities while preserving it on every other identity, in contrast to unlearning, which seeks to obtain a model indistinguishable from one retrained without the data to be unlearned. We do this in the open set, both across and within identities. Prior work on unlearning for face recognition models operates on closed-set scenarios. We define concrete and measurable success conditions and metrics for the open-set problem.
     \item We propose training and evaluation protocols on the WebFace4M~\cite{DBLP:conf/cvpr/ZhuHDY0CZYLD021} dataset that reflect a realistic deployment scenario, in which the face recognition model is required to adversarially forget a chosen identity on the basis of a subset of that identity's capture instances available to the system, and to have the resulting forgetting generalise to further capture instances of the same identity acquired independently of the forgetting.
     \item We define capture instances and restrict the forgetting problem to across-instance pairs, since a within-instance pair is already linkable in pixel space without a face recognition model.
     \item We propose three new face forgetting methods: a centroid-push variant of the dispersion loss, and two per-image-target methods (extended CosFace and an orthonormal frame) that map each forget image to its own near-orthogonal target. We compare these methods against four prior-art baselines: the pairwise and hard dispersion losses of Zakharov~\cite{DBLP:journals/corr/abs-2512-13317}, the negative-gradient method NegGrad+~\cite{DBLP:conf/nips/KurmanjiTHT23}, and the unsupervised method CURE~\cite{DBLP:journals/corr/abs-2509-19562}.
     \item We evaluate our methods on the proposed protocol on various forget scales and backbones. We report the score distributions of each partition, and the verification and identification rates at operating points of $10^{-4}$ and $10^{-2}$.
\end{enumerate}

The remainder of the paper is organised as follows. \Cref{sec:related} surveys the relevant literature, \Cref{sec:methodology} formalises forgetting and presents prior-art baselines and proposed methods, \Cref{sec:experiments} describes the protocol and the training setup, \Cref{sec:results} reports the empirical comparison of the methods on the proposed protocol, and \Cref{sec:conclusions} concludes the paper.

\section{Related Work}
\label{sec:related}

We review machine unlearning and its face-identity and generative variants, the methods that evaluate forgetting, and the related backdoor and template-inversion attacks.

\begin{table*}[t]
  \centering
  \caption{Closest related work on identity unlearning, compared along the axes relevant to adversarial forgetting on a face recognition backbone.
    The forgetting-test column reports how each work probes whether forgetting has succeeded.}
  \label{tab:related}
  \scriptsize
  \setlength{\tabcolsep}{3pt}
  \renewcommand{\arraystretch}{1.25}
  \newcolumntype{L}[1]{>{\raggedright\arraybackslash}p{#1}}
  \begin{tabular}{@{}L{2.3cm} L{5.9cm} L{2.0cm} L{3.4cm} L{3.6cm}@{}}
    \toprule
    Work & Approach / loss & Eval.\ unit & Forgetting test & Unlearning goal \\
    \midrule
    MUFAC~\cite{DBLP:journals/corr/abs-2311-02240}
      & Benchmark of fine-tuning and gradient-ascent unlearning baselines for face age and attribute classification
      & Single image
      & Loss-based membership-inference classifier
      & Forgotten images indistinguishable from unseen ones \\
    Nair et al.~\cite{NairKSA23_selective_unlearning_fr}
      & Fine-tuning
      & Single image
      & Single-probe verification
      & Selective accuracy reduction \\
    Hayes et al.~\cite{DBLP:journals/corr/abs-2403-01218}
      & Per-example membership-inference attack
      & Single sample
      & Membership inference
      & Match a from-scratch model \\
    Kim et al.~\cite{DBLP:journals/corr/abs-2503-06991}
      & Representation-level evaluation by feature similarity to the retrained model and nearest-neighbour transfer
      & Encoder features
      & Representation similarity, downstream transfer
      & Match a from-scratch model \\
    UMA~\cite{DBLP:journals/corr/abs-2504-14798}
      & Adversarial-query audit whose optimised input perturbations resurface forgotten outputs
      & Perturbed input
      & White-box, pre- and post-unlearning access
      & No adversarially recoverable knowledge \\
    Seo et al.~\cite{DBLP:conf/cvpr/SeoLLMP24}
      & Fine-tune a face generator so each identity's latent maps past the average face
      & Generated image
      & Re-synthesis from a latent code
      & Generator cannot synthesise the identity \\
    Nguyen et al.~\cite{DBLP:journals/corr/abs-2512-06562}
      & Per-identity surrogate latent with a continual-learning regulariser to retain other identities
      & Generated image
      & Re-synthesis of many identities
      & Generator cannot synthesise the identities \\
    \midrule
    \rowcolor{genericrow} NegGrad+~\cite{DBLP:conf/nips/KurmanjiTHT23}
      & Fine-tune the trained model on the retain and forget sets together, negating the gradient of the forget term
      & Single sample
      & Forget-set error, membership inference adapted from LiRA
      & Match a from-scratch model, per application \\
    \rowcolor{genericrow} CURE~\cite{DBLP:journals/corr/abs-2509-19562}
      & Feature erasure from K-means farthest-cluster pseudo-labels and a centroid-guided contrastive loss
      & Image, feature
      & Forget-set accuracy reduction, membership inference
      & Label-free erasure of the forget set \\
    \rowcolor{priorrow} Zakharov~\cite{DBLP:journals/corr/abs-2512-13317}
      & Repel same-identity embeddings apart on the sphere, with a hard-positive variant
      & Image, centroid
      & Closed-set retrieval (rank-1, mean average precision)
      & Make forgotten identities unretrievable \\
    \midrule
    \rowcolor{ourrow}\textbf{This paper}
      & Centroid push, an extended classifier head, and orthonormal-frame targets
      & Image, template
      & Open-set verification and identification
      & Adversarial forgetting \\
    \bottomrule
  \end{tabular}
\end{table*}

\textbf{Machine unlearning.}
Machine unlearning was introduced by Cao and Yang~\cite{DBLP:conf/sp/CaoY15} as a reformulation of the learning algorithm into a summation form, under which forgetting an individual training sample reduces to an update of the stored summands rather than retraining the model from scratch.
The exact-unlearning methods that followed aim at a model that is statistically indistinguishable from one retrained on the reduced training set, either via partitioned training that confines each sample to a small subset of shards~\cite{DBLP:conf/sp/BourtouleCCJTZL21}, or via closed-form parameter updates available only in specific hypothesis classes such as linear classifiers and $k$-means clustering~\cite{DBLP:conf/nips/GinartGVZ19,DBLP:conf/icml/GuoGHM20}.
Approximate methods, which are the only practical option for deep networks, modify the trained model directly through one of several geometrically distinct mechanisms, including closed-form influence-function updates~\cite{DBLP:conf/ndss/WarneckePWR23}, Fisher-importance-based selective parameter dampening~\cite{DBLP:conf/aaai/FosterSB24}, anchored gradient optimisation that penalises drift from the original parameters~\cite{DBLP:journals/corr/abs-2506-14515}, and theoretical bounds on the number of samples that can be deleted under convex losses~\cite{DBLP:conf/nips/SekhariAKS21}.
A separate strand uses Lipschitz regularisation to smooth forget outputs~\cite{DBLP:journals/corr/abs-2402-01401,DBLP:conf/wacv/KravetsN25}, in contrast to the use of the Lipschitz property as a lower bound on intra-identity cosine that we exploit later in \Cref{sec:methodology}.
The entirety of this literature operates under the train-from-scratch-equivalence interpretation of unlearning that we draw in the introduction, and is for that reason not directly comparable to the methods we propose under the selective-performance-removal interpretation we adopt as the operational target.

\textbf{Face identity unlearning.}
Face identity unlearning emerged as a distinct subproblem in which the model under attack is a face recognition model rather than a generic classifier, and the operational goal is to remove the verifiability of selected identities rather than the residual influence of an individual training sample.
On the discriminative side, the methods most comparable to ours operate on the embedding geometry of the face recognition model, with Zakharov~\cite{DBLP:journals/corr/abs-2512-13317} formalising the problem through a pairwise dispersion loss that repels same-identity embeddings on the unit sphere, together with a hard-positive mining variant. Shivam et al.\ (CURE)~\cite{DBLP:journals/corr/abs-2509-19562} take an unsupervised route in which K-means farthest-cluster pseudo-labels are combined with a centroid-guided contrastive objective and the results are summarised through an ``Unlearning Efficiency Score''.
Another benchmarking effort, from Choi and Na~\cite{DBLP:journals/corr/abs-2311-02240}, introduces the MUFAC and MUCAC benchmarks for instance-level personal-identity unlearning evaluated through face classification accuracy and a membership-inference classifier rather than through verification protocols, and is complemented by the fine-tuning procedure of Nair et al.~\cite{NairKSA23_selective_unlearning_fr} for the same setting.
On the generative side, Seo et al.~\cite{DBLP:conf/cvpr/SeoLLMP24} unlearn identities from a 3D-aware generative face model by remapping each erased identity to a target latent code obtained by extrapolating past the average-face latent, and Nguyen et al.~\cite{DBLP:journals/corr/abs-2512-06562} extend the construction to the multi-identity setting through a personalised surrogate latent per identity.
Of the lines surveyed in this section, this is the one to which our work is most directly comparable, and the discussion in the positioning paragraph below addresses it in detail.

\textbf{Concept erasure and (vision)-language-model unlearning.}
A related line of work is concept erasure, in which a concept, rather than an individual training sample, is unlearned from a generative model.
These image-domain methods differ in the stage at which they intervene. Gandikota et al.~\cite{DBLP:conf/iccv/GandikotaMFB23}, Kumari et al.~\cite{DBLP:conf/iccv/KumariZWS0Z23}, and the Forget-Me-Not procedure of Zhang et al.~\cite{DBLP:journals/corr/abs-2303-17591} act at fine-tuning time, the first by matching the conditional score function on the target concept to the unconditional one under negative guidance, the second by aligning the target distribution to that of an anchor concept that should remain available, and the third by suppressing the cross-attention of target tokens.
Of the remaining image-domain methods, Schramowski et al.~\cite{DBLP:conf/cvpr/SchramowskiBDK23} steer the diffusion sampler at inference time away from unsafe concept regions without modifying the underlying weights, Heng and Soh~\cite{DBLP:conf/nips/HengS23} reformulate the problem as continual learning toward a surrogate distribution to which the model should converge, and Fan et al.~\cite{DBLP:conf/iclr/FanLZ0W024} introduce saliency-weighted parameter updates that apply equally to classification and generative unlearning.
Closest to a general-purpose unlearning method among our baselines, Kurmanji et al.~\cite{DBLP:conf/nips/KurmanjiTHT23} combine gradient ascent on the forget set with a retain-side anchor for deep classifiers, the NegGrad+ procedure we adapt to the face recognition head.
The same selective-degradation framing appears in large language model unlearning at scale, with representative work including approximate unlearning of fictional-world content~\cite{DBLP:journals/corr/abs-2310-02238}, the TOFU benchmark of fictitious authors~\cite{DBLP:journals/corr/abs-2401-06121}, gradient ascent with retain-side regularisation in pretrained LLMs~\cite{DBLP:conf/acl/YaoCDNWCY24}, and a survey arguing for evaluation protocols beyond forget-set accuracy and perplexity~\cite{DBLP:journals/corr/abs-2402-08787}.

\textbf{Evaluation of unlearning.}
A growing body of work argues that current unlearning evaluations overstate the privacy or the forgetting they claim, and proposes stronger probes against which the unlearned model should be tested.
On the membership-inference side, Hayes et al.~\cite{DBLP:journals/corr/abs-2403-01218} demonstrate that the population U-MIAs commonly used in this literature miss vulnerable samples that a per-example U-MIA exposes, Kim et al.~\cite{DBLP:journals/corr/abs-2503-06991} replace logit-based metrics on small unlearning tasks with representation-level evaluation when the forget classes are semantically close to the downstream-task classes, and Du et al.~\cite{DBLP:journals/corr/abs-2406-13348} show that textual unlearning leaves recoverable residual structure and can instead expose the unlearned text.
On the verification side, Zhang et al.~\cite{DBLP:conf/icml/ZhangCSL24} find that current unlearning-verification mechanisms can be circumvented by an adversarial model provider that retains the relevant information while passing the checks, and Xu et al.~\cite{DBLP:journals/corr/abs-2410-10120} recover unlearned samples directly from the post-unlearning model parameters.
A further class of probes, such as UMA~\cite{DBLP:journals/corr/abs-2504-14798} on the discriminative side and the prompt-attack work of Jang et al.~\cite{DBLP:journals/corr/abs-2506-10236} and Yuan et al.~\cite{DBLP:journals/corr/abs-2408-10682} on the language-model side, crafts adversarial queries or prompts that expose residual knowledge in models claimed to have been unlearned.
Two additional results sit alongside these probes, with Tsiolakis et al.~\cite{DBLP:journals/corr/abs-2508-16150} reporting that some unlearning methods worsen the privacy of retained data while improving that of forgotten data, and Huang et al.~\cite{DBLP:journals/corr/abs-2410-09591} attacking the unlearning interface itself by submitting adversarial unlearning requests for data that the model never observed.
The broader shift in this literature, from passive single-probe evaluation toward probes that test whether forgetting holds beyond the samples operated on, is the one we adopt and make precise for face verification in the methodology and experimental sections that follow.

\textbf{Backdoors and template inversion.}
Adversarial forgetting is the opposite of the backdoor and template-inversion attacks studied on face recognition systems, which make the system accept a comparison it should reject, whereas adversarial forgetting makes it reject comparisons it would otherwise accept. The backdoor remains related nonetheless, since it tampers with a model already in service, which makes the detection of such tampering a concern common to backdooring and adversarial forgetting, whereas a template inversion leaves the model untouched and turns its stored templates against it.
Backdoor attacks engineer the model such that a chosen input is falsely accepted as a target, where the trigger may be a fixed training-data pattern~\cite{DBLP:journals/corr/abs-1708-06733,DBLP:journals/access/GuLDG19,DBLP:journals/corr/abs-1712-05526,DBLP:conf/ndss/LiuMALZW018}, a post-deployment poisoning of the template mechanism~\cite{DBLP:conf/eurosp/LovisottoEM20}, a physical-world artifact such as glasses or stickers~\cite{DBLP:conf/cvpr/WengerPBY0Z21}, or a facial-feature pattern~\cite{DBLP:journals/corr/abs-2006-11623}.
Template-inversion attacks produce the same accept response from the other direction and decode a stored embedding into a usable face image that the face recognition system then re-verifies~\cite{DBLP:journals/pami/MaiCYJ19,DBLP:conf/iccv/Otroshi-Shahreza23}.
In adversarial forgetting, by contrast, the defender is the one who modifies the model, and the input space is left untouched.
The two problems also share an evaluation weakness, since a backdoor that only fires under aggregation and an unlearning method that succeeds only on isolated probes would both pass single-input testing.

Among the works discussed in this section, Zakharov~\cite{DBLP:journals/corr/abs-2512-13317} and CURE~\cite{DBLP:journals/corr/abs-2509-19562} are closest in goal, and the evaluation-side line of Hayes et al.~\cite{DBLP:journals/corr/abs-2403-01218}, Kim et al.~\cite{DBLP:journals/corr/abs-2503-06991}, and UMA~\cite{DBLP:journals/corr/abs-2504-14798} closest in pushing unlearning evaluation toward stronger probes.
We differ in framing adversarial forgetting as a problem of matching score distributions, in which the distribution of mated forget comparison scores is brought onto the distribution of comparisons between unrelated identities, and in studying the problem in the open set with respect to the forget identity.
To our knowledge, ours is the only work to score forgetting at the calibrated operating point of a face recognition system, through one-to-one verification and open-set identification, rather than through the rank of a retrieved image or a membership-inference check. \Cref{tab:related} compares this to previous work.

\section{Methodology}
\label{sec:methodology}

\subsection{Motivation}
\label{ssec:motivation}

We are concerned with making a chosen set of identities ``unlinkable'' across the separate appearances in which they are recorded, while the model remains in service for the rest of the population. Unlinkability of this kind cannot be obtained by deleting the images of those identities from the training data, since a typical face recognition model still recognises identities it never observed in training. The embedding space must therefore be altered deliberately against the chosen identities themselves, which is what makes the forgetting adversarial.
For an identity the model has been tuned to forget, no two images of that person captured on separate occasions are to be matched to one another, so that, for example, an image taken from surveillance footage should not match the photograph held in that person's bona fide identity document.
On every other identity the model must behave as it did before the forgetting, such that the rates reported by the standard face recognition benchmarks stay close to their original values.

The problem concerns the images of an identity recorded on separate occasions, which differ in pose, illumination, and the camera that produced them.
Two images from a single capture can already be linked by an object tracker, since the face moves only by rotation and translation between them, and no face recognition model is needed to establish that one person recurs within one recording.
Such a pair is also close in pixel space, and the networks used for face recognition are Lipschitz continuous, so that the distance between the two embeddings is bounded by a constant multiple of the distance between the two images. Pushing those embeddings apart while their inputs remain near-identical is therefore difficult, since it requires a large increase in that constant.

\subsection{Problem statement}
\label{ssec:problem}

\textbf{Definitions and notation.} Let $\mathcal X$ be the set of aligned face images, and let $\mathcal Y$ be the finite set of identities present in it, with a labelling $\mathrm{id} : \mathcal X \rightarrow \mathcal Y$ that maps each image to its identity.
For an identity $y \in \mathcal Y$ we write $X_y = \mathrm{id}^{-1}(y) \subseteq \mathcal X$ for the set of images belonging to that identity.
We distinguish three pairwise disjoint subsets $\mathcal F, \mathcal R, \mathcal T \subseteq \mathcal Y$, which we call the \textit{forget set}, the \textit{retain set}, and the \textit{test set}, respectively.
The forget set $\mathcal F$ contains the identities to be adversarially forgotten, the retain set $\mathcal R$ those that must stay linkable, and the test set $\mathcal T$ untargeted identities that serve as a control on whether the rest of the population is left intact.
The face recognition model is trained on the retain identities $\mathcal R$, whereas the forget set $\mathcal F$ and the test set $\mathcal T$ are held out of its training, as we are dealing with an open-set problem.
The face recognition model is a map
\begin{equation}
  f^{\ast} : \mathcal X \rightarrow S^{d-1},
  \qquad f^{\ast}(x) = g^{\ast}(x) / \|g^{\ast}(x)\|_2,
\end{equation}
that takes an image to an $\Ltwo$-normalised embedding on the unit sphere $S^{d-1} \subset \mathbb R^d$, where $g^{\ast}$ is the unnormalised output of the backbone.
A verification system is fully specified by $f^{\ast}$ and a similarity threshold $\tau$ calibrated at a target false-match rate (FMR) on the non-mated comparisons of a development set $\mathcal D$ disjoint from $\mathcal R$, $\mathcal F$, and $\mathcal T$.
In our experiments, $f^{\ast}$ is a convolutional network~\cite{DBLP:conf/icpr/DutaL0020} or a Vision Transformer~\cite{DBLP:conf/iclr/DosovitskiyB0WZ21}.

\textbf{Linkability.} Two images $x_a, x_b \in \mathcal X$ are linked with respect to $f^{\ast}$ and $\tau$ if and only if $\langle f^{\ast}(x_a), f^{\ast}(x_b)\rangle \ge \tau$, and unlinked otherwise.
An identity stays linkable while some pair of its images is linked, since the system then assigns two of its appearances to one person, and it is this pairwise link that must be broken by adversarial forgetting.

\textbf{Identification.} The same model is also used to search a probe image against a gallery of enrolled identities rather than against a single claimed identity.
A forget identity is enrolled in that gallery alongside the retain identities, and forgetting in identification is the condition that a probe image of a forget identity, taken from an appearance that did not contribute to its enrolled template, is not identified as that identity.

\begin{figure}[t]
  \centering
  \subfloat[Within-instance pairs.]{%
    \begin{minipage}[t]{0.235\columnwidth}\centering
    \includegraphics[width=\linewidth]{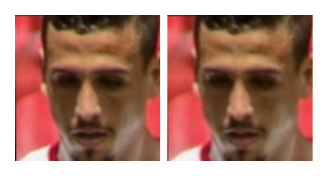}\\[-3pt]
    {\scriptsize $\delta{=}8,\,s{=}0.97$}%
  \end{minipage}\hfill
    \begin{minipage}[t]{0.235\columnwidth}\centering
    \includegraphics[width=\linewidth]{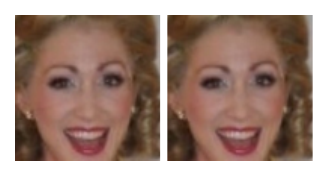}\\[-3pt]
    {\scriptsize $\delta{=}3,\,s{=}0.98$}%
  \end{minipage}\hfill
    \begin{minipage}[t]{0.235\columnwidth}\centering
    \includegraphics[width=\linewidth]{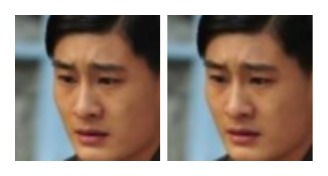}\\[-3pt]
    {\scriptsize $\delta{=}6,\,s{=}0.97$}%
  \end{minipage}\hfill
    \begin{minipage}[t]{0.235\columnwidth}\centering
    \includegraphics[width=\linewidth]{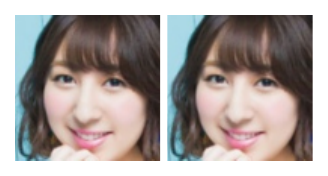}\\[-3pt]
    {\scriptsize $\delta{=}6,\,s{=}0.99$}%
  \end{minipage}%
    \label{fig:capture-within}}\\[4pt]
  \subfloat[Across-instance pairs.]{%
    \begin{minipage}[t]{0.235\columnwidth}\centering
    \includegraphics[width=\linewidth]{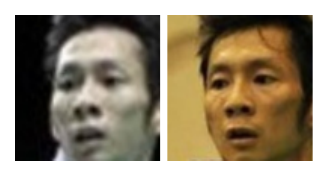}\\[-3pt]
    {\scriptsize $\delta{=}134,\,s{=}0.74$}%
  \end{minipage}\hfill
    \begin{minipage}[t]{0.235\columnwidth}\centering
    \includegraphics[width=\linewidth]{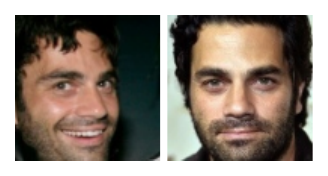}\\[-3pt]
    {\scriptsize $\delta{=}136,\,s{=}0.66$}%
  \end{minipage}\hfill
    \begin{minipage}[t]{0.235\columnwidth}\centering
    \includegraphics[width=\linewidth]{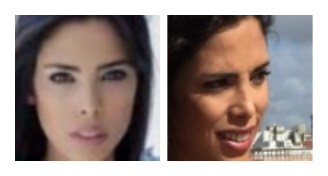}\\[-3pt]
    {\scriptsize $\delta{=}120,\,s{=}0.64$}%
  \end{minipage}\hfill
    \begin{minipage}[t]{0.235\columnwidth}\centering
    \includegraphics[width=\linewidth]{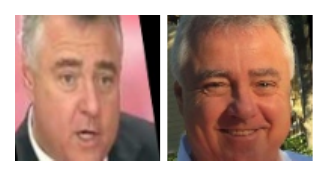}\\[-3pt]
    {\scriptsize $\delta{=}115,\,s{=}0.62$}%
  \end{minipage}%
    \label{fig:capture-across}}
  \caption{Example pairs of images from single forget identities, labelled with the pixel distance $\delta = \norm{x_a - x_b}_2$ and the embedding inner product $s = \langle f^{\ast}(x_a), f^{\ast}(x_b)\rangle$.}
  \label{fig:capture}
\end{figure}

\textbf{Capture instances and Lipschitz continuity.} The images in $X_y$ are rarely independent draws from an underlying appearance distribution of $y$, since many of them come from a single acquisition event such as one exposure, one video segment, or one burst.
We call such a set of images a \textit{capture instance}, within which the images differ only in low-level factors such as sensor noise, compression, and small temporal change, while images from different capture instances differ in pose, expression, illumination, and scene.
Two images from one capture instance form a \textit{within-instance} pair, and two images from different capture instances form an \textit{across-instance} pair, with examples of each shown in \Cref{fig:capture}.
An across-instance pair is linked only through the backbone of the face recognition model, and the system recognises a person across separate events through that link.
Face recognition backbones are Lipschitz continuous, and therefore there exists some Lipschitz constant $L$ bounding $\norm{f^{\ast}(x_a) - f^{\ast}(x_b)}_2 \le L \norm{x_a - x_b}_2$ for every pair of images.
We write $\varepsilon := x_a - x_b$ for the difference between two images and $\theta_{ab}$ for the angle between their embeddings, so that this constant gives the lower bound $\cos\theta_{ab} \ge 1 - \tfrac{L^2}{2}\|\varepsilon\|_2^{\,2}$, which depends on the weights rather than on the objective under which they were trained.
The similarity of a within-instance pair therefore stays as high under the altered model as under $f^{\ast}$.
Pushing such a pair apart works against the continuity of the architecture, and the link that would be broken is one that an object tracker already provides without recognition, so we confine adversarial forgetting to across-instance pairs.

\textbf{Defining forgetting.} Adversarial forgetting produces an altered model $f : \mathcal X \rightarrow S^{d-1}$ from $f^{\ast}$, under which the chosen identities become unlinkable across capture instances while the rest of the population remains linkable.

\begin{figure}[t]
  \centering
  \includegraphics[width=\columnwidth]{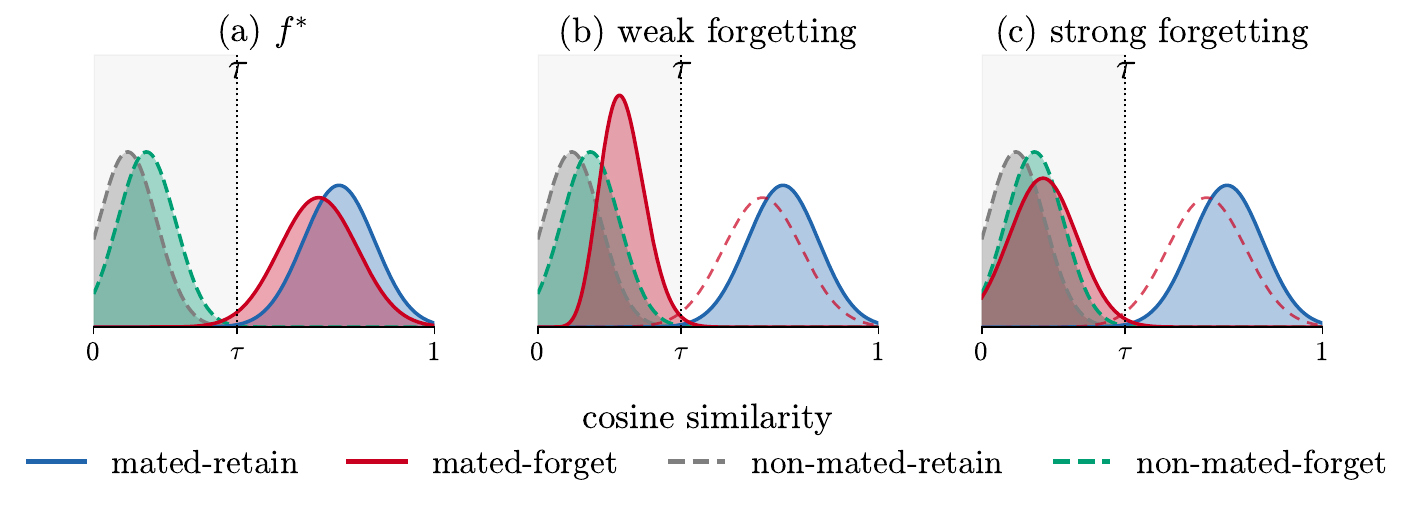}
  \caption{The four comparison-score distributions under the face recognition model $f^{\ast}$ (a), and under weak (b) and strong (c) forgetting. Under $f^{\ast}$ the mated-forget scores sit with the mated-retain scores above $\tau$, where images are said to be linked. Weak forgetting moves the mated-forget scores below $\tau$ but leaves them above the non-mated scores, whereas strong forgetting brings them onto the non-mated scores. The dashed red curve in (b) and (c) is the position of the mated-forget distribution under $f^{\ast}$.}
  \label{fig:distributions}
\end{figure}

\begin{figure*}[t]
  \centering
  \includegraphics[width=\textwidth]{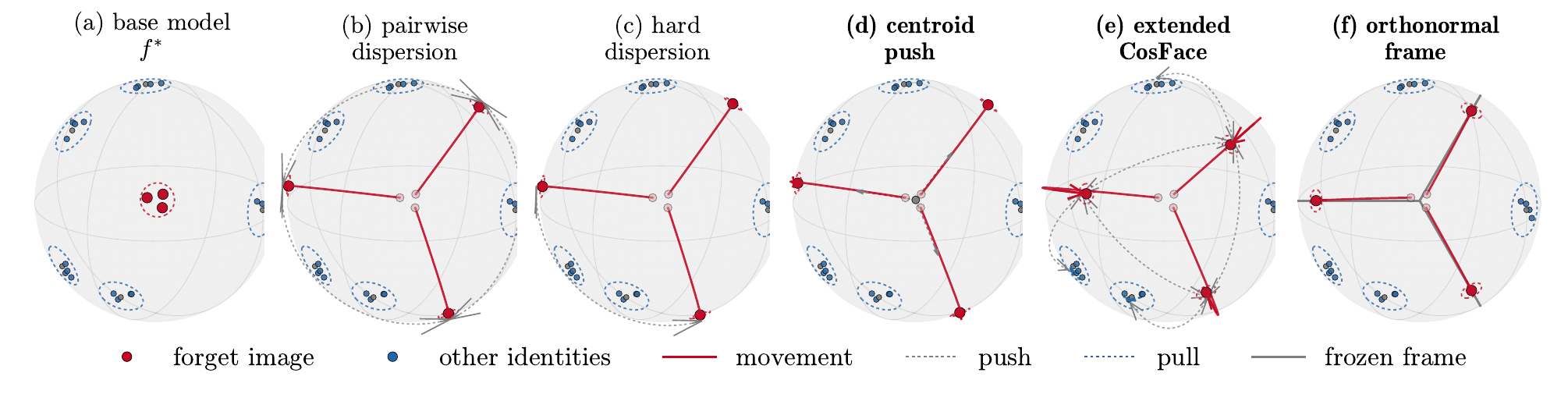}
  \caption{Mechanism of each forget loss on the unit sphere, with the forget positions obtained by minimising the real loss. The forget images start at the centre and the other identities sit on the periphery as dashed-circled clusters. The dispersion losses (b, c) and the centroid push (d) act on the embeddings with no classifier, extended CosFace (e) makes each image its own moving class centre under the margin head, and the orthonormal frame (f) anchors each image onto a frozen frame. Panels (d) to (f) are proposed in this work.}
  \label{fig:forget-mechanism}
\end{figure*}

The similarity scores of the mated and non-mated comparisons form four distributions, according to whether the comparison involves a forget identity, namely mated-retain, non-mated-retain, mated-forget, and non-mated-forget. The mated-forget distribution is that of the across-instance forget pairs of \Cref{fig:capture}.
Because $\tau$ is fixed on $\mathcal D$, it does not depend on the forget set or on any evaluated population. We now define two kinds of forgetting based on how mated-forget distributions behave under $f$ and $f^{\ast}$ with respect to $\tau$ and the non-mated retain distributions.

\textbf{Weak forgetting.} We say an identity $y \in \mathcal F$ is weakly forgotten under the altered model $f$ if and only if none of its across-instance pairs is linked under $f$, so that the system no longer assigns two of its appearances to one person.
The rest of the population should be left intact, so that the mated pairs of the retain identities stay linked and the non-mated pairs stay unlinked.
We write $\mathcal P_{\mathcal F}$ for the across-instance forget pairs over all $y \in \mathcal F$, on which the forget losses operate, $\mathcal P_{\mathcal R}$ for the mated pairs of the retain identities, and $\mathcal N$ for the non-mated pairs, and state the three conditions as
\begin{equation}
  \begin{aligned}
    \langle f(x_a), f(x_b)\rangle &\;<\; \tau, &\quad (x_a, x_b) &\in \mathcal P_{\mathcal F},\\
    \langle f(x_a), f(x_b)\rangle &\;\ge\; \tau, &\quad (x_a, x_b) &\in \mathcal P_{\mathcal R},\\
    \langle f(x_a), f(x_b)\rangle &\;<\; \tau, &\quad (x_a, x_b) &\in \mathcal N.
  \end{aligned}
  \label{eq:forget-goal}
\end{equation}
We seek an altered model $f$ that satisfies these conditions on as many pairs as possible.

\textbf{Strong forgetting.} The altered model $f$ should match the mated-forget distribution (which under $f^{\ast}$ coincides with mated-retain) to non-mated-retain (where the non-mated-forget distribution already sits), while the mated-retain and non-mated-retain distributions stay as close to their form under $f^{\ast}$ as possible.
We write $p_{\mathrm{mf}}$, $p_{\mathrm{mr}}$, and $p_{\mathrm{nr}}$ for the mated-forget, mated-retain, and non-mated-retain score distributions under $f$, and $p^{\ast}$ for the same distributions under $f^{\ast}$, and express strong forgetting as the optimisation
\begin{equation}
  \min_{f} \; \lambda_1 D(p_{\mathrm{mf}} \| p_{\mathrm{nr}})
  + \lambda_2 D(p_{\mathrm{mr}} \| p^{\ast}_{\mathrm{mr}})
  + \lambda_3 D(p_{\mathrm{nr}} \| p^{\ast}_{\mathrm{nr}}),
  \label{eq:strong-forgetting}
\end{equation}
where $D$ is a distributional distance such as a Wasserstein distance, and the weights satisfy $\lambda_1, \lambda_2, \lambda_3 \ge 0$ and $\lambda_1 + \lambda_2 + \lambda_3 = 1$, so that they trade the forgetting against the preservation of the retain distributions.

Under weak forgetting, the mated-forget comparison scores fall below $\tau$ yet stay above the non-mated comparison scores, so the forget identities are unlinkable at the operating point while their comparisons still form a sub-population that the non-mated comparisons do not produce and that can only come from the forget identities. This information may then be used to link forget identities subject to weak forgetting. Under strong forgetting, no such sub-population remains, since the mated-forget distribution coincides with non-mated-retain, and the comparisons of a forget identity become indistinguishable from comparisons between unrelated identities, which is the sense in which we call the identities strongly forgotten.
\Cref{fig:distributions} shows the distributions of the face recognition model and the effects of weak and strong forgetting.

\subsection{Forget losses}
\label{ssec:losses}

The forget losses act on the across-instance forget pairs $\mathcal P_{\mathcal F}$ and degrade their similarity while preserving the behaviour of $f^{\ast}$ on the retain and test pairs. They follow one of two broad geometric mechanisms, a dispersion of a forget identity's images across the sphere or a mapping of each image onto its own approximately-orthogonal target. \Cref{fig:forget-mechanism} illustrates these losses, with one sphere per loss.
Each loss adds a forget term to the original CosFace retain loss, evaluated through the CosFace head $W^{\ast}$, warm-started from the face recognition model and fine-tuned jointly with the backbone,
\begin{equation}
  \label{eq:composite}
  \mathcal L(f) \;=\; \lambda_r\, \mathcal L_{\text{retain}}(f) \;+\; \lambda_f\, \mathcal L_{\text{forget}}(f),
\end{equation}
with $(\lambda_r, \lambda_f)$ setting the trade-off.
\paragraph{Prior methods.}
We adapt to our open-set verification and identification protocols four methods previously proposed for identity forgetting and machine unlearning, which we report as baselines.
The pairwise dispersion of Zakharov~\cite{DBLP:journals/corr/abs-2512-13317} penalises the cosine similarity of every across-instance forget pair through a hinge at a margin $m$,
\begin{equation}
  \label{eq:pd}
  \mathcal L_{\text{forget}}^{\text{PD}}(f) \;=\; \tfrac{1}{|\mathcal P_{\mathcal F}|} \sum_{(a, b) \in \mathcal P_{\mathcal F}} \max\!\bigl(0,\, m + f(x_a)^{\!\top} f(x_b)\bigr),
\end{equation}
and its hard-positive variant, the hard dispersion, keeps only the largest within-identity similarity of each image,
\begin{equation}
  \label{eq:hd}
  \mathcal L_{\text{forget}}^{\text{HD}}(f) \;=\; \tfrac{1}{N} \sum_{a} \max\!\Bigl(0,\, m + \max_{b\,:\,(a,b)\in\mathcal P_{\mathcal F}} f(x_a)^{\!\top} f(x_b)\Bigr),
\end{equation}
where $N$ is the number of forget images. \Cref{eq:pd} and \Cref{eq:hd} both act on the embedding geometry with no classifier head.
NegGrad+ of Kurmanji et al.~\cite{DBLP:conf/nips/KurmanjiTHT23} increases the forget cross-entropy, evaluated through one fake-class row per forget identity added to the head, while decreasing the retain cross-entropy.
CURE of Shivam et al.~\cite{DBLP:journals/corr/abs-2509-19562} repels each forget embedding from a frozen copy of $f^{\ast}$ towards the farthest-cluster pseudo-labels of a $K$-means partition, and distils the retain embeddings to hold them in place.
These are the objectives the prior methods compute.

In addition to these prior methods, we propose three losses of our own, which are defined as follows.

\paragraph{Centroid push.}
The centroid-push loss penalises each embedding's similarity to its identity's mean direction $\bar c_y(f) = \mathrm{norm}\bigl(\sum_{x \in X_y} f(x)\bigr)$, recomputed at each step or epoch,
\begin{equation}
  \mathcal L_{\text{forget}}^{\text{CP}}(f) \;=\; \tfrac{1}{|\mathcal F|}\!\sum_{y \in \mathcal F} \tfrac{1}{|X_y|}\!\sum_{x \in X_y}\!\bigl(1 + f(x)^{\!\top} \bar c_y(f)\bigr).
\end{equation}
Repelling every embedding from its identity's centroid disperses them at a cost linear rather than quadratic in $|X_y|$, the fixed point placing each embedding antipodal to that centroid.
It gives the same weak forgetting as pairwise dispersion, but more cheaply, except that the centroid is recomputed from the embeddings being pushed and so shifts as they move.

\paragraph{Extended CosFace with per-image fake-class rows.}
Extended CosFace augments the head $W^{\ast} \in \mathbb R^{|\mathcal R| \times d}$, warm-started from the face recognition model, whose rows we call the retain rows, with one row per forget image, so that each image $x_i$ has a fake class $c_i$ with a row $W_{c_i} \in S^{d-1}$ drawn at initialisation. It then applies the CosFace loss at scale $s_f$ and margin $m_f$ over the union of rows $W$.
The margin-softmax probability of target row $t$ among an arbitrary set of unit-norm rows $V$, instantiated with $W$, at scale $s_f$ and margin $m_f$ is
\begin{equation}
  \label{eq:marginsoftmax}
  \pi_t(x; V) \;=\; \frac{\exp\!\bigl(s_f(\langle f(x), V_t\rangle - m_f)\bigr)}{\sum_k \exp\!\bigl(s_f(\langle f(x), V_k\rangle - m_f\,\mathbf 1[k=t])\bigr)},
\end{equation}
and the extended-CosFace loss is
\begin{equation}
  \mathcal L_{\text{forget}}^{\text{EC}}(f) \;=\; -\sum_i \log \pi_{c_i}(x_i; W).
\end{equation}
Treating the images of one identity as separate classes makes each same-identity pair a non-mated comparison for the loss, and the angular margin spreads their embeddings onto near-orthogonal directions where similarities concentrate near zero.
The loss aims to achieve strong forgetting through the emergent geometric mechanisms of CosFace, which has a tendency to place different identities on almost orthogonal axes, as evidenced by the typical score distribution of non-mated trials for face recognition models trained with CosFace.

\paragraph{Orthonormal-frame targets.}
The orthonormal-frame loss replaces the random rows with a fixed set of targets $\{U_i\} \subset S^{d-1}$ built before training by greedy furthest-point sampling. Each target is chosen from random candidates to minimise its largest cosine similarity to the retain rows and the targets already placed, and it aligns each forget embedding with its own target directly,
\begin{equation}
  \mathcal L_{\text{forget}}^{\text{OF}}(f) \;=\; \tfrac{1}{N} \sum_i \bigl(1 - \langle f(x_i), U_i\rangle\bigr).
\end{equation}
The frame is almost-orthogonal, since the targets outnumber the orthogonal directions of $S^{d-1}$, and we refer to it as \textit{orthonormal} throughout.
Mapping each forget image's embedding onto its own target places the two embeddings of one identity in distinct directions, near-orthogonal to each other and to the retain rows, with similarity near zero.
This supplies the near-orthogonal arrangement of the embeddings by construction rather than as an emergent property of training, and therefore produces strong forgetting directly.

\section{Experimental Setup}
\label{sec:experiments}

This section specifies the dataset and the face recognition model, the adversarial forgetting methods and their training, the partitions, the verification and identification protocols, and the comparisons from which the metrics are computed.

\begin{table}[!tb]
  \centering
  \caption{Identity counts and mated comparison counts for the $S_1$ and $S_2$ protocols, with $1{,}000{,}000$ non-mated pairs sampled per group. The tr$\to$ev column counts the comparisons of the averaged forget-train enrolment against single forget-eval probes.}
  \label{tab:protocol}
  \small
  \setlength{\tabcolsep}{5pt}
  \resizebox{\columnwidth}{!}{%
  \begin{tabular}{@{}l ccccc@{}}
    \toprule
    Identity counts & $\mathcal F$ & $\mathcal R$ & $\mathcal T$ & Distr.\ & $\mathcal D$ \\
    \cmidrule(lr){1-1}\cmidrule(lr){2-6}
    $S_1$ & $1{,}000$ & $1{,}000$ & $1{,}000$ & $50{,}000$ & $35{,}850$ \\
    $S_2$ & $5{,}000$ & $1{,}000$ & $1{,}000$ & $50{,}000$ & $35{,}850$ \\
    \addlinespace[7pt]
    Mated comparisons & $\mathcal R$ & $\mathcal T$ & $\mathcal F$ ev--ev & $\mathcal F$ tr--tr & $\mathcal F$ tr$\to$ev \\
    \cmidrule(lr){1-1}\cmidrule(lr){2-6}
    $S_1$ & $10{,}488$ & $10{,}645$ & $33{,}917$ & $578{,}928$ & $6{,}289$ \\
    $S_2$ & $10{,}488$ & $10{,}645$ & $161{,}133$ & $2{,}760{,}850$ & $30{,}934$ \\
    \bottomrule
  \end{tabular}}
\end{table}

\subsection{Dataset and face recognition model}
\label{ssec:backbone}

We build the evaluation on WebFace4M~\cite{DBLP:conf/cvpr/ZhuHDY0CZYLD021}, a face dataset of $205{,}990$ identities and about $4.2$ million images.
The face recognition model is trained on $100{,}000$ of these identities, which form the retain set $\mathcal R$.
The remaining $105{,}990$ identities are never observed during training, and from them we draw the forget set $\mathcal F$, the test set $\mathcal T$, the development set $\mathcal D$, and the distractors of the identification gallery.

The backbone is an iResNet-34~\cite{DBLP:conf/icpr/DutaL0020} with a CosFace head~\cite{DBLP:conf/cvpr/WangWZJGZL018} of scale $64$ and margin $0.4$, as in the InsightFace recipe of Deng et al.~\cite{DBLP:conf/cvpr/DengGXZ19}, and the embedding dimension is $d = 512$.
Training runs for 40 epochs by stochastic gradient descent at a batch size of $128$, a learning rate of $0.1$ on a polynomial schedule, a momentum of $0.9$, and a weight decay of $5\times10^{-4}$.
This trained backbone is the face recognition model $f^{\ast}$ from which every altered model is fine-tuned.

\subsection{Forgetting methods and training}
\label{ssec:training}

Every altered model is fine-tuned from $f^{\ast}$ under the composite objective of \Cref{eq:composite}, whose retain term is computed over all $100{,}000$ retain identities.
Each forget batch has $64$ images drawn by an identity-balanced sampler over $\mathcal F$.

Before training, we deduplicate the images of each forget identity by keeping a single image from each capture instance and removing every other image of that instance. Two images belong to the same capture instance when their intensity-normalised $\Ltwo$ distance in pixel space falls below $30$, on the scale $\delta$ of \Cref{fig:capture}.
Deduplication is needed because the forget losses operate on the pairs of a forget identity's images, and without it those pairs would include within-instance pairs, whose inputs are Lipschitz-close and whose similarity therefore stays high however long the training runs.
The forget term on such a pair cannot be reduced, so it contributes a persistent gradient that destabilises the training.

Every altered model is fine-tuned for 40 epochs, and we evaluate the final checkpoint in each case rather than selecting a checkpoint by validation.
Under the two dispersion losses and the centroid push, the fine-tuning runs on a single GPU at a learning rate of $10^{-3}$ with $\lambda_f = 10\lambda_r$, and the hinge margin of the dispersion losses is $m = 0.35$.
Under extended CosFace and the orthonormal frame, it runs on four GPUs at a learning rate of $5\times10^{-3}$ with $\lambda_f = \lambda_r$, the forget margin of extended CosFace is $m_f = 0.65$ at the head scale $s_f = 64$, and the forget term of the frame is scaled by a factor of $100$.
Under NegGrad+ and CURE it runs at a learning rate of $10^{-3}$, such that the training budget is held fixed across every method rather than taken from each published configuration.
In the tables we abbreviate the pairwise and hard dispersion as PD and HD, NegGrad+ as NG+, the centroid push as CP, extended CosFace as EC, and the orthonormal frame as OF.

\subsection{Partitions}
\label{ssec:partitions}

The operating points are fixed on the development set $\mathcal D$, whose images serve as the non-mated probes of the identification protocol.
An identity is admitted to the retain control, the test set $\mathcal T$, or the development set only if it has at least four images.
The forget and test identities are both untrained, so the comparisons between test identities are the reference against which the forget comparisons are read.

We split each forget identity's deduplicated images into a forget-train partition $\mathcal F_{\mathrm{tr}}$ holding $80\%$ of them, on which the forget losses operate, and a held-out forget-eval partition $\mathcal F_{\mathrm{ev}}$ holding the remaining $20\%$.
We measure through this split whether forgetting on a subset of an identity's images generalises to the identity itself.
An identity left with too few deduplicated images to fill both partitions contributes only forget-train comparisons.
Because deduplication precedes the split, both partitions hold images from distinct capture instances, so their across-instance pairs become the forget pairs $\mathcal P_{\mathcal F}$ of \Cref{ssec:problem}.
We consider two forget sets of different scales, denoted $S_1$ and $S_2$, such that $S_1$ contains 1,000 identities (1\% of the training population) and $S_2$ contains 5,000 (5\%), with $S_1 \subset S_2$.

\subsection{Protocols}
\label{ssec:protocol}

In verification, a probe image is compared against the enrolled template of a single claimed identity.
The retain control and the test set $\mathcal T$ hold 1,000 identities each, and each identity is enrolled as one averaged template over the first half of its images and probed by the second half.
The retain control is a 1,000-identity sample of $\mathcal R$, written $\mathcal R$ in the tables, and is held to 1,000 only to balance the protocol against the forget set.
A forget identity is instead enrolled from each of its images on its own, in either partition, and additionally as one averaged template over its forget-train images, and is probed by single images from either partition, so that a mated forget comparison pairs two capture instances of the same identity.
\Cref{tab:protocol} reports the identity and comparison counts of the protocols, which differ between $S_1$ and $S_2$ only in the forget set, since the retain control, the test set $\mathcal T$, the distractors, and the development set $\mathcal D$ are drawn once with a shared seed and reused at both scales.
The threshold $\tau$ is calibrated on the non-mated comparisons of the development set $\mathcal D$, which excludes forget identities, at the operating points $10^{-4}$ and $10^{-2}$.

Each evaluated model is also used in open-set identification, where a probe is searched against the whole gallery of averaged templates.
The gallery holds the retain, test, and forget identities together with 50,000 additional untrained identities as distractors. A probe is identified when its highest-scoring gallery template belongs to its own identity and that score exceeds the threshold.
The forget identities are searched under two conditions that differ in how the gallery template is built.
A template built from the forget-train images models a database assembled before the forgetting optimisation, and a template built from the forget-eval images models re-enrolment after the forgetting.
At the same operating points the open-set threshold is calibrated on the development set $\mathcal D$, from which none of the identities is enrolled in the gallery.

\subsection{Comparisons and metrics}
\label{ssec:metrics}

\Cref{ssec:problem} names the comparison-score distributions by whether a comparison is mated and whether it involves a forget identity, and here the second axis is refined into the retain, test, and forget partitions.
The mated comparisons of each partition are accordingly mated-retain, mated-test, and mated-forget, and the comparisons between two different identities of $\mathcal R$ and of $\mathcal T$ are non-mated-retain and non-mated-test.
We take the non-mated-test distribution $p_{\mathrm{nt}}$ rather than the non-mated-retain distribution of \Cref{eq:strong-forgetting} as the reference for strong forgetting, since the test identities are untrained and are never targeted by the forgetting.

We group the mated forget comparisons by the partition from which the enrolment and the probe originate, into the forget-train comparisons ($\mathcal F_{\mathrm{tr}}$--$\mathcal F_{\mathrm{tr}}$), the forget-eval comparisons ($\mathcal F_{\mathrm{ev}}$--$\mathcal F_{\mathrm{ev}}$), and the comparisons of a forget-train enrolment against a forget-eval probe ($\mathcal F_{\mathrm{tr}}$--$\mathcal F_{\mathrm{ev}}$).
As with the gallery conditions of \Cref{ssec:protocol}, each group corresponds to a state of the database, which held both images before the forgetting optimisation, held only the enrolment, the probe being acquired after the forgetting, or held neither.
The forget losses operate only on the forget-train images, so the forget-eval comparisons measure whether the forgetting generalises to images on which the losses never operated.

In verification, the threshold is set at a target FMR, the rate at which non-mated comparisons are accepted. We report the true-match rate (TMR), the rate at which mated comparisons are accepted, equal to one minus the false-non-match rate (FNMR).
In identification, the operating point is set at a target false-positive identification rate (FPIR), the rate at which a non-mated probe is identified. We report the true-positive identification rate (TPIR), the rate at which a mated probe is correctly identified.
The tables of \Cref{sec:results} report each rate per partition and per method at $10^{-4}$ and $10^{-2}$, with the backbone comparison at $10^{-4}$.

The desired direction of these rates differs by partition, since on the retain set $\mathcal R$ and the test set $\mathcal T$ the TMR and the TPIR stay high when the population is preserved, whereas on the forget partitions $\mathcal F_{\mathrm{tr}}$ and $\mathcal F_{\mathrm{ev}}$ they decrease as the identities are forgotten.
The tables mark that direction on each partition.

For the forget identities, we also report the cross-forget FMR, computed over the non-mated-forget comparisons, which pair two different forget identities.
This rate stays at the FMR at which the threshold was calibrated when no forget identity is matched to another, whereas a higher rate marks that distinct forget identities stay linkable.

\section{Results and Discussion}
\label{sec:results}

We evaluate the proposed methods and four prior-art baselines on the $S_1$ and $S_2$ protocols against $f^{\ast}$.
Each group of methods carries its own tint in the tables, with the face recognition model in grey, the generic unlearning baselines in olive, the domain-specific prior art in light olive, and our methods in light grey.

\subsection{Findings}
\label{ssec:results-findings}

The main findings of our paper are as follows.
\begin{enumerate}
  \item\label{fnd:mechanisms} The effects of the forgetting methods can be classified into two distinct categories, which we call \textit{confusion} and \textit{manifold hiding}, associated with weak and strong forgetting respectively.
  The dispersion and centroid-push losses are \textit{confusers}, whereas the extended-CosFace and orthonormal-frame losses are \textit{manifold hiders} (\Cref{ssec:results-geometry}).
  \item\label{fnd:generic} The generic unlearning baselines are not competitive with the domain-specific methods under the open-set protocols.
  NegGrad+ leaves the forget rates at the level of the face recognition model, and the training under CURE produces degenerate models in which the retain identities are no longer identified (\Cref{ssec:results-generic}).
  \item\label{fnd:strong} Strong forgetting is best achieved with the orthonormal frame, and only where the comparison involves at least one image seen by the altered model during the forgetting.
  There the mated-forget scores fall onto the non-mated distribution almost perfectly, whereas between two unseen images of a forget identity they approach it yet remain distinguishable (\Cref{ssec:results-strong}).
  \item\label{fnd:generalise} Forgetting an identity's unseen images is harder than forgetting the seen images.
  Weak forgetting on the unseen images nonetheless remains achievable to some extent for both the confusers and the manifold hiders (\Cref{ssec:results-generalisation}).
  \item\label{fnd:averaging} Feature averaging partly recovers a forgotten identity.
  Averaging image embeddings into an enrolment template undoes part of the forgetting for both the confusers and the manifold hiders (\Cref{ssec:results-generalisation}).
  \item\label{fnd:footprint} Each method leaves a signature \textit{footprint} on the score distributions of the retain and test partitions.
  The footprint differs by method, and its shape lets an adversary recognise that the model has been altered (\Cref{ssec:results-footprint}).
  \item\label{fnd:backbone} The forgetting carries to other backbones.
  The general trends are similar across the backbones we test, and the type of backbone influences mainly the performance of the altered model on the retain identities (\Cref{ssec:results-scale}).
  \item\label{fnd:verdict} No single method is best on every requirement, and the choice follows the priority.
  The orthonormal frame gives the stronger forgetting, whereas the hard dispersion gives the smaller cost and footprint.
  We maintain the recommendation at larger forget scales (\Cref{ssec:results-cost}).
\end{enumerate}
The rest of this section presents the experimental results and discusses each finding in detail.

\subsection{Performance of Generic Unlearning Baselines}
\label{ssec:results-generic}

The two baselines adapted from generic unlearning, NegGrad+ and CURE, are not competitive with the domain-specific methods. NegGrad+ leaves the TMR on the forget identities comparable to the rate under $f^{\ast}$, at $0.998$ against $0.912$ on the forget-train images (\Cref{tab:results}), i.e., there is practically no forgetting. CURE degrades the retain verification and identification performance, at a retain TMR of $0.895$ and a retain TPIR of $0.137$ at $10^{-4}$ (\Cref{tab:results}). Furthermore, on $S_1$ every domain-specific method gives a lower forget TPIR than CURE, and every one except the centroid push a lower forget TMR.

Both baselines produce score distributions that are either degenerate (CURE) or barely moved (NegGrad+). \Cref{tab:kl} reports two relevant quantities. The first is the Wasserstein-1 distance $W_1(p_{\bullet}, p_{\mathrm{nt}})$ between the distribution of mated comparison scores of each forget partition and the distribution of non-mated comparison scores of the test set, which measures how closely a method moves the mated-forget scores onto the scores of unrelated identities. It corresponds to the first term of \Cref{eq:strong-forgetting} with the non-mated-retain reference replaced by the non-mated-test distribution. The second quantity reported in \Cref{tab:kl}, which we call the \textit{deformation}, is the Wasserstein-1 distance between a retain score distribution of the altered model and the same distribution of $f^{\ast}$, which measures how far the forgetting has moved the scores of the retained identities from their shape before the optimisation. We report two deformations, one for the mated and one for the non-mated retain comparisons. CURE displaces the mated and the non-mated retain distributions far from their shape in $f^{\ast}$, at deformations of $0.313$ and $0.943$ on $S_1$ against at most $0.101$ for every other method (\Cref{tab:kl}), and the resulting model no longer identifies most of the retain identities at $\mathrm{FPIR}=10^{-4}$ (\Cref{tab:results}). NegGrad+ does not move the mated-forget distributions towards the non-mated scores, and places the mated-forget-train distribution further from the non-mated-test distribution than under $f^{\ast}$, at $0.968$ against $0.480$ (\Cref{tab:kl}). The forget comparisons therefore become more distinctive after the unlearning rather than less.

We therefore focus on the other losses in the rest of the paper, since the behaviour of these baselines offers no further insight into the forgetting mechanisms. Their rates stay in the tables, because the claim that they are not competitive can only be made from their reported performance.

\begin{table}[!tb]
  \centering
  \caption{Wasserstein-1 distances between score distributions, in units of the cosine similarity. The deformation columns give the distance of the mated and non-mated retain score distributions from their shape under $f^{\ast}$ (\Cref{ssec:results-generic}). Best values within a margin of $0.002$ are in bold.}
  \label{tab:kl}
  \footnotesize
  \setlength{\tabcolsep}{2.5pt}
  \begin{tabular}{@{}l cc cc cc@{}}
    \toprule
    & \multicolumn{2}{c}{$W_1(p_{\bullet}, p_{\mathrm{nt}})$, $S_1$} & \multicolumn{2}{c}{Deformation, $S_1$} & \multicolumn{2}{c}{Deformation, $S_2$} \\
    \cmidrule(lr){2-3}\cmidrule(lr){4-5}\cmidrule(lr){6-7}
    Method & $\mathcal F_{\mathrm{tr}}$ & $\mathcal F_{\mathrm{ev}}$ & Mated & Non-mated & Mated & Non-mated \\
    \midrule
    \rowcolor{refrow} $f^{\ast}$ & $0.480$ & $0.480$ & $0.000$ & $0.000$ & $0.000$ & $0.000$ \\
    \cmidrule(l){1-7}
    \rowcolor{genericrow} CURE~\cite{DBLP:journals/corr/abs-2509-19562} & $0.026$ & $0.025$ & $0.313$ & $0.943$ & $0.244$ & $0.739$ \\
    \rowcolor{genericrow} NG+~\cite{DBLP:conf/nips/KurmanjiTHT23} & $0.968$ & $0.740$ & $0.027$ & $\mathbf{0.002}$ & $\mathbf{0.010}$ & $0.003$ \\
    \cmidrule(l){1-7}
    \rowcolor{priorrow} PD~\cite{DBLP:journals/corr/abs-2512-13317} & $0.162$ & $0.153$ & $0.020$ & $\mathbf{0.001}$ & $0.013$ & $\mathbf{0.000}$ \\
    \rowcolor{priorrow} HD~\cite{DBLP:journals/corr/abs-2512-13317} & $0.100$ & $\mathbf{0.103}$ & $\mathbf{0.014}$ & $\mathbf{0.001}$ & $0.018$ & $\mathbf{0.001}$ \\
    \cmidrule(l){1-7}
    \rowcolor{ourrow} CP & $0.435$ & $0.370$ & $0.025$ & $\mathbf{0.002}$ & $0.026$ & $\mathbf{0.001}$ \\
    \rowcolor{ourrow} EC & $0.058$ & $0.155$ & $\mathbf{0.015}$ & $\mathbf{0.003}$ & $0.037$ & $0.005$ \\
    \rowcolor{ourrow} OF & $\mathbf{0.006}$ & $0.110$ & $0.101$ & $\mathbf{0.001}$ & $0.238$ & $0.006$ \\
    \bottomrule
  \end{tabular}
\end{table}

\subsection{Confusion and Manifold Hiding}
\label{ssec:results-geometry}

\begin{figure*}[!t]
  \centering
  \includegraphics[width=\textwidth]{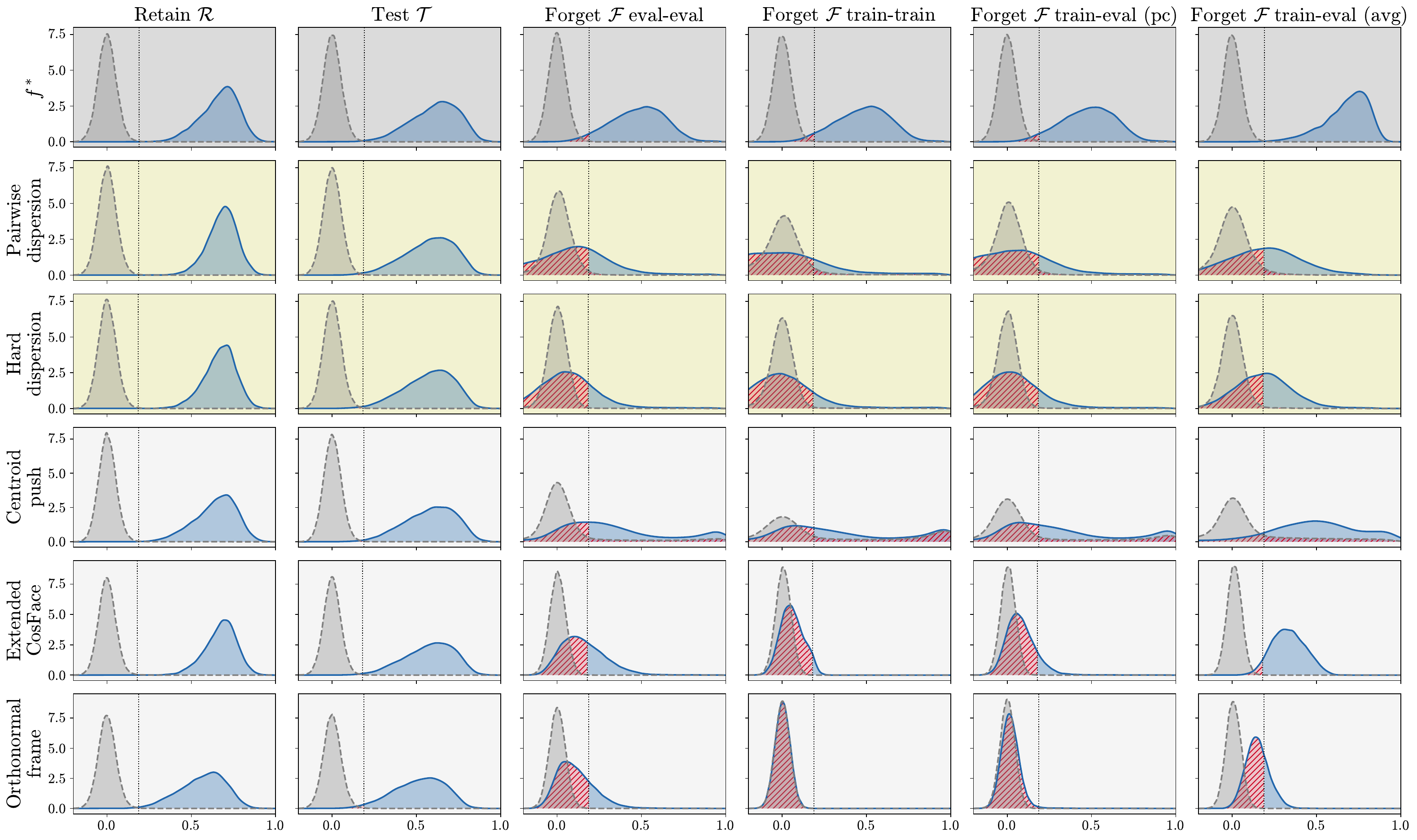}
  \caption{Distributions of mated (solid) and non-mated (dashed) comparison scores for the face recognition model $f^{\ast}$ and five forgetting methods. The columns are the retain set $\mathcal R$, the test set $\mathcal T$, the forget-eval and forget-train mated comparisons, and a forget-train enrolment against a forget-eval probe taken per image and as an averaged template. The threshold $\tau$ is calibrated on the non-mated non-forget comparisons at $\mathrm{FMR}=10^{-3}$.}
  \label{fig:dist}
\end{figure*}

Every method reshapes the distribution of the mated-forget comparison scores, and we identify two modes of reshaping. \Cref{fig:dist} reports the distributions of mated and non-mated comparison scores for the face recognition model and five forgetting methods, one row per model and one column per comparison group.

The two dispersion losses and centroid push spread the mated-forget scores into a broad, diffuse distribution that overlaps with the non-mated distribution and extends past it towards higher similarity scores.
The dispersion losses optimise the mated-forget distribution to have a heavy right tail, whereas the centroid push moves most of its mass above the non-mated scores.
Extended CosFace and the orthonormal frame instead concentrate the forget-train scores ($\mathcal F_{\mathrm{tr}}$) into a narrow, thin-tailed peak on the non-mated distribution that leaves almost no mass above $\tau$.
We call the first behaviour \textit{confusion} and the second \textit{manifold hiding}, and the two shapes are what separate the confusers from the manifold hiders of Finding~\ref{fnd:mechanisms}.

Our results indicate that confusion corresponds to the weak forgetting and manifold hiding to the strong forgetting defined in~\Cref{ssec:problem}.
Manifold hiders move the distribution of mated-forget comparison scores onto the distribution of non-mated comparison scores of the same comparison group. The mated-forget comparisons thus become nearly indistinguishable from comparisons between unrelated identities.

A small $W_1(p_{\bullet}, p_{\mathrm{nt}})$ on a forget partition in \Cref{tab:kl} means mated-forget comparisons coincide with the non-mated-test distribution, which is a signature of manifold hiders, whereas a larger value marks the residual cluster of confusion.
The two manifold hiders give the two smallest $W_1$ distances to the non-mated-test distribution and the three confusers the three largest, from $0.006$ for the orthonormal frame to $0.435$ for the centroid push (\Cref{tab:kl}) on forget-train comparisons.

The confusers also make distinct forget identities linkable to one another, i.e., a comparison between images of two distinct forget identities is scored higher than $\tau$ more often than a comparison between two unrelated identities.
On the forget-train images of $S_1$ every confuser raises the rate above the FMR at which the threshold was calibrated, most strongly the centroid push at $0.3$ against a calibrated $0.01$. In contrast, the cross-forget FMR stays at or below the calibrated rate for manifold hiders.
The cross-forget FMR is also higher on the forget-train than on the forget-eval images for every confuser, so the linkability is strongest for images seen by forget losses.

The distinctions and the characteristics of the confusers and the manifold hiders are similar across the remaining experiments, with the rates varying and with the single exception of the hard dispersion, which approaches manifold hiding on the transformer backbones (\Cref{ssec:results-scale}).
The dispersion losses are the best of the confusers, and the orthonormal frame is the better of the two manifold hiders.

\subsection{Limits of Strong Forgetting in Manifold Hiders}
\label{ssec:results-strong}

Only manifold hiders approach strong forgetting, and only on train partitions.
The orthonormal frame moves the distribution of mated-forget comparison scores between two forget-train images onto the non-mated-test distribution (the forget-train column of \Cref{fig:dist}), and the two distributions become nearly identical.
The $W_1$ distance between the two is $0.006$, against $0.48$ under the face recognition model, and the forget-train TMR in \Cref{tab:results} is zero at $\mathrm{FMR}=10^{-4}$ and $0.002$ at $10^{-2}$ on $S_1$.
The strong forgetting extends to the comparisons between a forget-train and a forget-eval image, at a $W_1$ distance of $0.02$ to the non-mated retain distribution (the per-image enrolment column).
Extended CosFace follows the same pattern on the seen images yet stops short of strong forgetting, at a forget-train TMR of $0.185$ at $\mathrm{FMR}=10^{-2}$.

The strong forgetting of manifold hiders does not generalise to the unseen images of a forget identity.
Between two forget-eval images the mated-forget distribution moves below the threshold instead of onto the non-mated distribution (the forget-eval column of \Cref{fig:dist}).
The orthonormal frame leaves a tail past $\tau$, the $W_1$ distance to the non-mated-test distribution increases from $0.006$ to $0.11$, and the distance for extended CosFace increases from $0.058$ to $0.155$ (\Cref{tab:kl}).
For example, at $\mathrm{FMR}=10^{-4}$ the forget-eval TMR is $0.106$ for the orthonormal frame and $0.232$ for extended CosFace, against forget-train TMRs of zero and $0.001$ (\Cref{tab:results}).
Therefore, we can characterise the behaviour of manifold hider-altered models on unseen crops of forget-identities as weak forgetting.

In comparison, dispersion losses achieve similar levels of forgetting on the seen and the unseen images in the form of weak forgetting.
Their $W_1$ distances to the non-mated-test distribution are nearly equal on the two partitions, $0.100$ against $0.103$ for the hard dispersion and $0.162$ against $0.153$ for the pairwise dispersion (\Cref{tab:kl}), although the lower tail of their mated-forget distribution extends to similarities below the smallest non-mated retain scores.
The manifold hiders trade this uniformity for the near-perfect forgetting of the seen images.
This limit of strong forgetting under manifold-hiding is therefore a limit of generalisation within the identity, which we examine in the next subsection.

\begin{table*}[!t]
  \centering
  \caption{Verification TMR@FMR and open-set identification TPIR@FPIR of each partition at $10^{-4}$ and $10^{-2}$, on the $S_1$ ($1\%$) and $S_2$ ($5\%$) forget sets of the iResNet-34. $\uparrow$ marks a partition where a higher rate is better and $\downarrow$ one where a lower rate is better. Best values per forget set within a margin of $0.002$ are in bold, excluding the face recognition model $f^{\ast}$. The $\mathcal F{\times}\mathcal F$ rows report the cross-forget FMR, the false-match rate between two distinct forget identities at the threshold of each operating point. \Cref{fig:det} reports the forget rates across the whole operating range.}
  \label{tab:results}
  \footnotesize
  \setlength{\tabcolsep}{3.1pt}
  \setlength{\aboverulesep}{0pt}
  \setlength{\belowrulesep}{0pt}
  \renewcommand{\arraystretch}{1.03}
  \begin{tabular}{@{}ll >{\columncolor{refrow}}c>{\columncolor{genericrow}}c>{\columncolor{genericrow}}c|>{\columncolor{priorrow}}c>{\columncolor{priorrow}}c>{\columncolor{ourrow}}c>{\columncolor{ourrow}}c>{\columncolor{ourrow}}c|>{\columncolor{refrow}}c>{\columncolor{genericrow}}c>{\columncolor{genericrow}}c|>{\columncolor{priorrow}}c>{\columncolor{priorrow}}c>{\columncolor{ourrow}}c>{\columncolor{ourrow}}c>{\columncolor{ourrow}}c@{}}
    \toprule
    & & \multicolumn{8}{c|}{$S_1$} & \multicolumn{8}{c}{$S_2$} \\
    \cmidrule(lr){3-18}
    & & $f^{\ast}$ & CURE & NG+ & PD & HD & CP & EC & OF & $f^{\ast}$ & CURE & NG+ & PD & HD & CP & EC & OF \\
    \midrule
    \multicolumn{18}{c}{\textbf{Verification (TMR@FMR)}} \\
    \midrule

    \multirow{4}{*}{$10^{-4}$} & $\mathcal R\,\uparrow$ & $1.000$ & $0.895$ & $\mathbf{1.000}$ & $\mathbf{1.000}$ & $\mathbf{1.000}$ & $\mathbf{0.998}$ & $\mathbf{1.000}$ & $0.985$ & $1.000$ & $0.859$ & $\mathbf{1.000}$ & $\mathbf{1.000}$ & $\mathbf{1.000}$ & $\mathbf{0.998}$ & $\mathbf{1.000}$ & $0.912$ \\
    & $\mathcal T\,\uparrow$ & $0.988$ & $0.869$ & $0.978$ & $0.978$ & $\mathbf{0.980}$ & $0.979$ & $\mathbf{0.982}$ & $0.952$ & $0.988$ & $0.832$ & $\mathbf{0.977}$ & $0.973$ & $\mathbf{0.977}$ & $\mathbf{0.976}$ & $\mathbf{0.978}$ & $0.825$ \\
    & $\mathcal F_{\mathrm{tr}}\,\downarrow$ & $0.912$ & $0.351$ & $0.998$ & $0.165$ & $0.097$ & $0.423$ & $\mathbf{0.001}$ & $\mathbf{0.000}$ & $0.918$ & $0.316$ & $0.974$ & $0.167$ & $0.059$ & $0.432$ & $0.153$ & $\mathbf{0.009}$ \\
    & $\mathcal F_{\mathrm{ev}}\,\downarrow$ & $0.912$ & $0.256$ & $0.829$ & $0.216$ & $0.128$ & $0.485$ & $0.232$ & $\mathbf{0.106}$ & $0.921$ & $0.312$ & $0.806$ & $0.266$ & $0.118$ & $0.548$ & $0.258$ & $\mathbf{0.042}$ \\
    \cmidrule(lr){1-18}
    \multirow{4}{*}{$10^{-2}$} & $\mathcal R\,\uparrow$ & $1.000$ & $0.996$ & $\mathbf{1.000}$ & $\mathbf{1.000}$ & $\mathbf{1.000}$ & $\mathbf{1.000}$ & $\mathbf{1.000}$ & $\mathbf{1.000}$ & $1.000$ & $0.982$ & $\mathbf{1.000}$ & $\mathbf{1.000}$ & $\mathbf{1.000}$ & $\mathbf{1.000}$ & $\mathbf{1.000}$ & $0.987$ \\
    & $\mathcal T\,\uparrow$ & $0.999$ & $0.992$ & $\mathbf{0.998}$ & $\mathbf{0.998}$ & $\mathbf{0.998}$ & $\mathbf{0.998}$ & $\mathbf{0.998}$ & $0.995$ & $0.999$ & $0.972$ & $\mathbf{0.998}$ & $\mathbf{0.996}$ & $\mathbf{0.997}$ & $\mathbf{0.997}$ & $\mathbf{0.998}$ & $0.959$ \\
    & $\mathcal F_{\mathrm{tr}}\,\downarrow$ & $0.986$ & $0.735$ & $0.999$ & $0.294$ & $0.226$ & $0.549$ & $0.185$ & $\mathbf{0.002}$ & $0.987$ & $0.803$ & $0.985$ & $0.314$ & $0.181$ & $0.632$ & $0.505$ & $\mathbf{0.071}$ \\
    & $\mathcal F_{\mathrm{ev}}\,\downarrow$ & $0.985$ & $0.617$ & $0.860$ & $0.431$ & $\mathbf{0.330}$ & $0.666$ & $0.548$ & $0.380$ & $0.988$ & $0.801$ & $0.886$ & $0.498$ & $0.337$ & $0.764$ & $0.605$ & $\mathbf{0.220}$ \\
    \midrule
    \multicolumn{18}{c}{\textbf{Cross-forget verification (FMR)}} \\
    \midrule
    \multirow{2}{*}{$10^{-4}$} & $\mathcal F_{\mathrm{tr}}{\times}\mathcal F_{\mathrm{tr}}\,\downarrow$ & $0.000$ & $\mathbf{0.000}$ & $0.013$ & $0.036$ & $0.012$ & $0.265$ & $\mathbf{0.000}$ & $\mathbf{0.000}$ & $0.000$ & $\mathbf{0.000}$ & $0.019$ & $0.054$ & $0.005$ & $\mathbf{0.000}$ & $\mathbf{0.000}$ & $\mathbf{0.000}$ \\
    & $\mathcal F_{\mathrm{ev}}{\times}\mathcal F_{\mathrm{ev}}\,\downarrow$ & $0.000$ & $\mathbf{0.000}$ & $0.009$ & $0.005$ & $\mathbf{0.002}$ & $0.106$ & $\mathbf{0.000}$ & $\mathbf{0.000}$ & $0.000$ & $\mathbf{0.000}$ & $0.011$ & $0.011$ & $\mathbf{0.000}$ & $\mathbf{0.000}$ & $\mathbf{0.000}$ & $\mathbf{0.000}$ \\
    \cmidrule(lr){1-18}
    \multirow{2}{*}{$10^{-2}$} & $\mathcal F_{\mathrm{tr}}{\times}\mathcal F_{\mathrm{tr}}\,\downarrow$ & $0.010$ & $\mathbf{0.001}$ & $0.036$ & $0.104$ & $0.028$ & $0.300$ & $0.004$ & $\mathbf{0.002}$ & $0.010$ & $\mathbf{0.001}$ & $0.064$ & $0.089$ & $0.019$ & $0.012$ & $0.007$ & $0.006$ \\
    & $\mathcal F_{\mathrm{ev}}{\times}\mathcal F_{\mathrm{ev}}\,\downarrow$ & $0.010$ & $\mathbf{0.000}$ & $0.026$ & $0.047$ & $0.014$ & $0.140$ & $0.006$ & $0.004$ & $0.009$ & $\mathbf{0.001}$ & $0.042$ & $0.034$ & $0.010$ & $0.011$ & $0.007$ & $0.006$ \\
    \midrule
    \multicolumn{18}{c}{\textbf{Identification (TPIR@FPIR)}} \\
    \midrule

    \multirow{4}{*}{$10^{-4}$} & $\mathcal R\,\uparrow$ & $0.902$ & $0.137$ & $0.015$ & $0.946$ & $0.933$ & $0.000$ & $\mathbf{0.951}$ & $0.694$ & $0.900$ & $0.000$ & $0.000$ & $0.185$ & $0.912$ & $0.817$ & $\mathbf{0.925}$ & $0.447$ \\
    & $\mathcal T\,\uparrow$ & $0.744$ & $0.145$ & $0.005$ & $0.622$ & $0.649$ & $0.000$ & $\mathbf{0.683}$ & $0.553$ & $0.741$ & $0.000$ & $0.000$ & $0.080$ & $0.580$ & $\mathbf{0.623}$ & $0.608$ & $0.295$ \\
    & $\mathcal F_{\mathrm{tr}}\,\downarrow$ & $0.850$ & $0.309$ & $0.510$ & $0.062$ & $0.049$ & $0.017$ & $0.054$ & $\mathbf{0.000}$ & $0.854$ & $\mathbf{0.000}$ & $0.387$ & $0.012$ & $0.016$ & $0.243$ & $0.278$ & $0.031$ \\
    & $\mathcal F_{\mathrm{ev}}\,\downarrow$ & $0.676$ & $0.105$ & $0.374$ & $0.041$ & $0.019$ & $\mathbf{0.002}$ & $0.073$ & $0.020$ & $0.686$ & $\mathbf{0.000}$ & $0.204$ & $\mathbf{0.002}$ & $0.014$ & $0.222$ & $0.076$ & $0.010$ \\
    \cmidrule(lr){1-18}
    \multirow{4}{*}{$10^{-2}$} & $\mathcal R\,\uparrow$ & $0.979$ & $0.594$ & $\mathbf{0.996}$ & $\mathbf{0.997}$ & $0.992$ & $0.943$ & $\mathbf{0.995}$ & $0.860$ & $0.978$ & $0.264$ & $0.971$ & $\mathbf{0.990}$ & $\mathbf{0.989}$ & $0.943$ & $\mathbf{0.990}$ & $0.646$ \\
    & $\mathcal T\,\uparrow$ & $0.889$ & $0.558$ & $0.824$ & $0.834$ & $0.838$ & $0.830$ & $\mathbf{0.852}$ & $0.745$ & $0.886$ & $0.263$ & $0.752$ & $0.800$ & $0.815$ & $0.814$ & $\mathbf{0.819}$ & $0.485$ \\
    & $\mathcal F_{\mathrm{tr}}\,\downarrow$ & $0.933$ & $0.673$ & $0.715$ & $0.094$ & $0.087$ & $0.317$ & $0.172$ & $\mathbf{0.003}$ & $0.936$ & $0.472$ & $0.775$ & $0.059$ & $\mathbf{0.044}$ & $0.410$ & $0.492$ & $0.102$ \\
    & $\mathcal F_{\mathrm{ev}}\,\downarrow$ & $0.843$ & $0.420$ & $0.609$ & $0.115$ & $\mathbf{0.048}$ & $0.298$ & $0.256$ & $0.090$ & $0.848$ & $0.203$ & $0.548$ & $0.123$ & $0.049$ & $0.433$ & $0.271$ & $\mathbf{0.029}$ \\
    \bottomrule
  \end{tabular}
\end{table*}

\subsection{Generalisation Within an Identity and Re-linking}
\label{ssec:results-generalisation}

\begin{figure*}[!tb]
  \centering
  \begin{minipage}[t]{0.49\textwidth}
    \vspace{0pt}
    \centering
    \subfloat[Verification]{\includegraphics[width=0.49\linewidth]{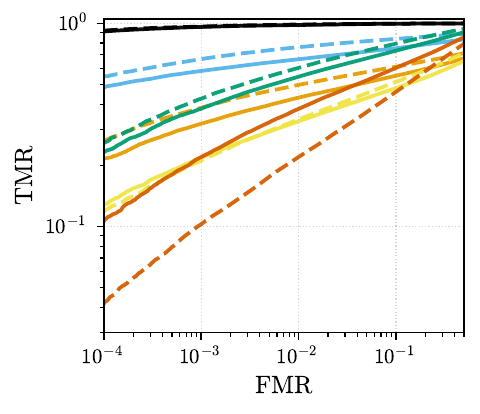}}\hfill
    \subfloat[Identification]{\includegraphics[width=0.49\linewidth]{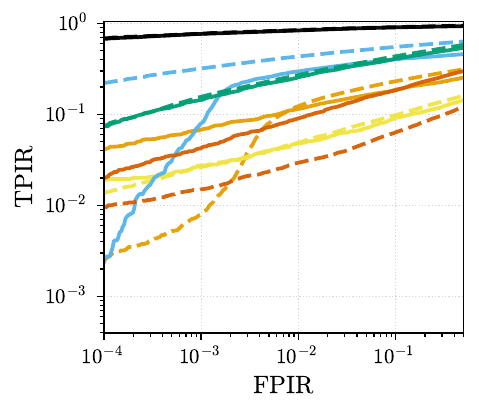}}\\
    \includegraphics[width=0.9\linewidth]{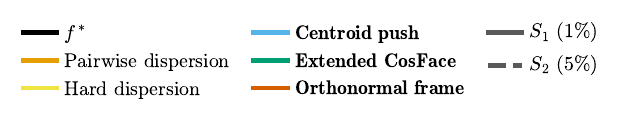}
    \caption{Forget-identity verification (a) and open-set identification (b) across the operating range, on the $S_1$ (solid) and $S_2$ (dashed) forget sets. NegGrad+ and CURE are excluded (\Cref{ssec:results-generic}).}
    \label{fig:det}
  \end{minipage}\hfill
  \begin{minipage}[t]{0.49\textwidth}
    \vspace{0pt}
    \centering
    \includegraphics[width=\linewidth]{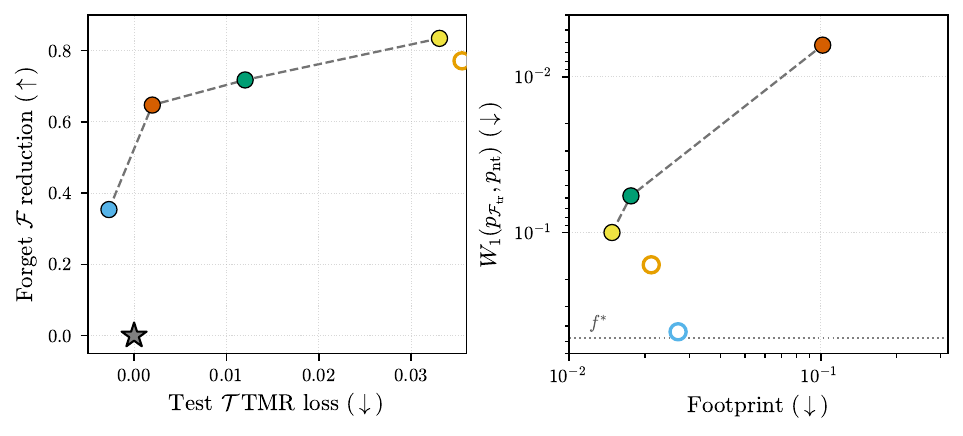}\\[3pt]
    \includegraphics[width=0.9\linewidth]{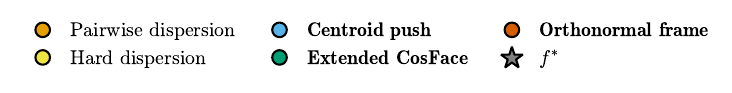}
    \caption{Weak and strong forgetting against their costs on the $S_1$ forget set, as the forget-eval TMR reduction against the test TMR loss in verification (left), both at the retain-comparison equal-error threshold of each model, and $W_1(p_{\mathcal F_{\mathrm{tr}}}, p_{\mathrm{nt}})$ against the footprint (right). A filled marker is on the Pareto front of its panel, and the dotted line indicates the distance under $f^{\ast}$.}
    \label{fig:pareto}
  \end{minipage}
\end{figure*}

Forgetting an identity's unseen images is harder than forgetting the seen images (Finding~\ref{fnd:generalise}).
Wherever a forget-train image enters the comparison, the forgetting is easier than on the unseen images for most methods, and its degree ranges from weak for the confusers to strong for the manifold hiders (\Cref{fig:dist}).
Between two forget-eval images the degree of the forgetting decreases for the confusers as well.
The forget-eval TMR stays above the forget-train TMR at every operating point (the $\mathcal F_{\mathrm{tr}}$ and $\mathcal F_{\mathrm{ev}}$ rows of \Cref{tab:results}).
Below an FMR of $10^{-3}$ the orthonormal frame gives the lowest forget-eval TMR, and above it the hard dispersion gives the lowest rate (\Cref{fig:det}(a)).
Every method we compare makes forget identities nearly unidentifiable on $S_1$, whether the gallery template is built from forget-train or forget-eval images (\Cref{tab:results}).

We also consider an oracle with access to linkage information for the individual images of the forget identities, such as a database of enrolled images assembled before the forgetting optimisation.
The oracle knows which images belong to the same forget identity, a linkage the altered model no longer provides, since the forgetting pushes the mated forget comparisons below the threshold.
We then build a template for each forgotten identity by averaging the embeddings of its forget-train images, an enrolment we could not assemble without the oracle-provided linkage.
We probe the averaged template with a forget-eval image and plot the distribution of mated-forget scores in the last column of \Cref{fig:dist}.
Against the averaged enrolment, part of the distribution of mated-forget comparison scores moves back above $\tau$ for every method (the last two columns of \Cref{fig:dist}).
The mass above $\tau$ increases most for extended CosFace, from $0.12$ against the single-image enrolment to $0.95$ against the averaged enrolment, and the averaging undoes even weak forgetting.
The mass increase above $\tau$ is the smallest for the orthonormal frame, from $0.01$ to $0.27$.
External linkage thus brings part of the identity information back for the confusers and the manifold hiders alike (Finding~\ref{fnd:averaging}).
Individual forget images are difficult to link to one another, as the forget TMR and TPIR in \Cref{tab:results} show, yet once images are linked, the averaged template makes a new image of the identity easier to link.
Forgetting an identity with these methods therefore requires the deletion of previously linked templates from legacy galleries as well, on top of modifying the model.

\subsection{The Footprint and the Cost to the Retained Identities}
\label{ssec:results-footprint}

Furthermore, the cost that the forgetting imposes on the retained identities varies by method (Finding~\ref{fnd:footprint}).
The performance under weak forgetting can be checked on the verification and identification tasks in \Cref{tab:results}, which reports the TMR and the TPIR of each partition at FMR/FPIR of $10^{-4}$ and $10^{-2}$.
In verification the cost is negligible, and on $S_1$ the retain TMR stays within $0.015$ of the face recognition model's rate for every method except CURE.
In identification the cost is larger, and it arises through two mechanisms: one that raises the open-set threshold and one that broadens the lower tail of the distribution of mated retain comparison scores.
The centroid push incurs the severest cost through the first mechanism, since the forget embeddings that it disperses stay enrolled in the gallery and produce high scores for unknown probes, which raises the open-set threshold until, at $\mathrm{FPIR}=10^{-4}$, no mated retain probe is accepted (\Cref{fig:det}(b)).
The orthonormal frame maps its forget embeddings onto fixed targets, which produce no high scores for unknown probes, so the cross-forget FMR stays at or below the calibrated rate and the threshold is unaffected.
The cost of the orthonormal frame arrives through the second mechanism, in which the mated retain scores broaden towards lower similarities and the retain probes that fall below the threshold are no longer identified.
Neither the raised threshold nor the broadened lower tail arises for the dispersion losses and extended CosFace, since the dispersion losses disperse the forget embeddings more mildly, at cross-forget FMRs of at most $0.036$ against $0.265$ for the centroid push, and extended CosFace concentrates the forget embeddings instead of dispersing them.
The retain TPIR therefore stays at or above the face recognition model's rate for the dispersion losses and extended CosFace (\Cref{tab:results}).

The deformations of \Cref{ssec:results-generic} measure the second mechanism directly, and the deformation columns of \Cref{tab:kl} report them per method and forget set.
On $S_1$ the deformation of the distribution of non-mated retain comparison scores does not exceed $0.003$ for any method.
The deformation of the distribution of mated retain comparison scores stays between $0.01$ and $0.03$ for every method except the orthonormal frame, which raises it to $0.10$, and the broadening of the lower tail of the mated retain scores is visible in the retain column of \Cref{fig:dist}.
We define the \textit{footprint} as the sum of the two deformations, which summarises in one quantity how far an altered model has moved the retain score distributions.
Among the confusers and the manifold hiders, the footprint is smallest for the hard dispersion, at $0.015$ on $S_1$, and largest for the orthonormal frame, at $0.10$ on $S_1$ and $0.24$ on $S_2$.
The footprint measures the size of this movement and not its direction, which differs between the two groups.
The dispersion losses and extended CosFace concentrate the distribution of mated retain comparison scores, whose standard deviation decreases from $0.110$ for the face recognition model to between $0.087$ and $0.095$, and whose $0.1\%$ quantile increases from $0.291$ to at least $0.339$. Their footprint therefore measures an improvement in the separability of the retain identities rather than a cost.
The orthonormal frame widens the same distribution, to an estimated standard deviation of $0.134$ and a $0.1\%$ quantile of $0.144$, and the retain TMR decreases from $1.000$ to $0.985$ at $\mathrm{FMR}=10^{-4}$ (\Cref{tab:results}).
The orthonormal frame incurs that cost in exchange for the closest match of the mated-forget scores to the non-mated scores wherever an image seen during training is included in the protocol.
In a system whose enrolled images are known before the forgetting optimisation, and on which the forget losses therefore operate, the orthonormal frame gives a $W_1$ distance of $0.006$ from the non-mated-test distribution between two forget-train images and of $0.02$ from the non-mated retain distribution between a forget-train and a forget-eval image, against $0.100$ and $0.094$ for the hard dispersion.
\Cref{fig:pareto} reports the strong forgetting of each method against its footprint.

An adversary who knows how the mated and non-mated comparison scores of the face recognition model are distributed can measure the same two distributions on a model in service, and can compute from the discrepancy the likelihood that the model has been trained to behave differently.
A large footprint therefore indicates that the model in service has strayed from the ordinary training lifecycle, and this detectability is a cost of the forgetting that is paid whatever recognition performance the altered model preserves.

\subsection{Scale and Backbones}
\label{ssec:results-scale}

\begin{table*}[!t]
  \centering
  \caption{Backbone comparison on $S_1$ at an FMR or FPIR of $10^{-4}$, over the partitions of \Cref{tab:results} and the three forget comparison groups. Best values per column within $0.002$ are in bold, excluding $f^{\ast}$. NegGrad+ and CURE are excluded (\Cref{ssec:results-generic}).}
  \label{tab:backbones}
  \footnotesize
  \setlength{\tabcolsep}{4pt}
  \setlength{\aboverulesep}{0pt}
  \setlength{\belowrulesep}{0pt}
  \renewcommand{\arraystretch}{1.03}
  \begin{tabular}{@{}l ccccc ccccc ccccc@{}}
    \multicolumn{16}{c}{\textbf{Verification (TMR@FMR)}} \\
    \toprule
    & \multicolumn{5}{c}{iResNet-34} & \multicolumn{5}{c}{ViT-S} & \multicolumn{5}{c}{ViT-B} \\
    \cmidrule(lr){2-6}\cmidrule(lr){7-11}\cmidrule(lr){12-16}
    Method & $\mathcal R\,\uparrow$ & $\mathcal T\,\uparrow$ & \shortstack{$\mathcal F_{\mathrm{tr}}$-\\$\mathcal F_{\mathrm{tr}}\,\downarrow$} & \shortstack{$\mathcal F_{\mathrm{tr}}$-\\$\mathcal F_{\mathrm{ev}}\,\downarrow$} & \shortstack{$\mathcal F_{\mathrm{ev}}$-\\$\mathcal F_{\mathrm{ev}}\,\downarrow$} & $\mathcal R\,\uparrow$ & $\mathcal T\,\uparrow$ & \shortstack{$\mathcal F_{\mathrm{tr}}$-\\$\mathcal F_{\mathrm{tr}}\,\downarrow$} & \shortstack{$\mathcal F_{\mathrm{tr}}$-\\$\mathcal F_{\mathrm{ev}}\,\downarrow$} & \shortstack{$\mathcal F_{\mathrm{ev}}$-\\$\mathcal F_{\mathrm{ev}}\,\downarrow$} & $\mathcal R\,\uparrow$ & $\mathcal T\,\uparrow$ & \shortstack{$\mathcal F_{\mathrm{tr}}$-\\$\mathcal F_{\mathrm{tr}}\,\downarrow$} & \shortstack{$\mathcal F_{\mathrm{tr}}$-\\$\mathcal F_{\mathrm{ev}}\,\downarrow$} & \shortstack{$\mathcal F_{\mathrm{ev}}$-\\$\mathcal F_{\mathrm{ev}}\,\downarrow$} \\
    \midrule
    \rowcolor{refrow} $f^{\ast}$ & $1.000$ & $0.988$ & $0.912$ & $0.911$ & $0.912$ & $1.000$ & $0.982$ & $0.889$ & $0.882$ & $0.884$ & $1.000$ & $0.987$ & $0.905$ & $0.901$ & $0.900$ \\
    \cmidrule(l){1-16}
    \rowcolor{priorrow} PD~\cite{DBLP:journals/corr/abs-2512-13317} & $\mathbf{1.000}$ & $0.978$ & $0.165$ & $0.170$ & $0.216$ & $\mathbf{1.000}$ & $0.964$ & $0.156$ & $0.166$ & $0.225$ & $\mathbf{1.000}$ & $0.970$ & $0.233$ & $0.226$ & $0.246$ \\
    \rowcolor{priorrow} HD~\cite{DBLP:journals/corr/abs-2512-13317} & $\mathbf{1.000}$ & $\mathbf{0.980}$ & $0.097$ & $0.103$ & $0.128$ & $\mathbf{1.000}$ & $0.960$ & $0.038$ & $0.052$ & $\mathbf{0.090}$ & $\mathbf{1.000}$ & $0.965$ & $0.029$ & $0.039$ & $\mathbf{0.055}$ \\
    \cmidrule(l){1-16}
    \rowcolor{ourrow} CP & $\mathbf{0.998}$ & $0.979$ & $0.423$ & $0.429$ & $0.485$ & $0.981$ & $0.936$ & $0.205$ & $0.216$ & $0.260$ & $0.933$ & $0.909$ & $0.368$ & $0.404$ & $0.450$ \\
    \rowcolor{ourrow} EC & $\mathbf{1.000}$ & $\mathbf{0.982}$ & $\mathbf{0.001}$ & $0.044$ & $0.232$ & $\mathbf{1.000}$ & $0.973$ & $0.015$ & $0.098$ & $0.147$ & $\mathbf{1.000}$ & $0.981$ & $0.012$ & $0.110$ & $0.180$ \\
    \rowcolor{ourrow} OF & $0.985$ & $0.952$ & $\mathbf{0.000}$ & $\mathbf{0.005}$ & $\mathbf{0.106}$ & $\mathbf{1.000}$ & $\mathbf{0.980}$ & $\mathbf{0.000}$ & $\mathbf{0.013}$ & $0.097$ & $\mathbf{1.000}$ & $\mathbf{0.986}$ & $\mathbf{0.000}$ & $\mathbf{0.022}$ & $0.096$ \\
    \bottomrule
  \end{tabular}

  \vspace{4pt}
  \begin{tabular}{@{}l cccc cccc cccc@{}}
    \multicolumn{13}{c}{\textbf{Identification (TPIR@FPIR)}} \\
    \toprule
    & \multicolumn{4}{c}{iResNet-34} & \multicolumn{4}{c}{ViT-S} & \multicolumn{4}{c}{ViT-B} \\
    \cmidrule(lr){2-5}\cmidrule(lr){6-9}\cmidrule(lr){10-13}
    Method & $\mathcal R\,\uparrow$ & $\mathcal T\,\uparrow$ & $\mathcal F_{\mathrm{tr}}\,\downarrow$ & $\mathcal F_{\mathrm{ev}}\,\downarrow$ & $\mathcal R\,\uparrow$ & $\mathcal T\,\uparrow$ & $\mathcal F_{\mathrm{tr}}\,\downarrow$ & $\mathcal F_{\mathrm{ev}}\,\downarrow$ & $\mathcal R\,\uparrow$ & $\mathcal T\,\uparrow$ & $\mathcal F_{\mathrm{tr}}\,\downarrow$ & $\mathcal F_{\mathrm{ev}}\,\downarrow$ \\
    \midrule
    \rowcolor{refrow} $f^{\ast}$ & $0.902$ & $0.744$ & $0.850$ & $0.676$ & $0.999$ & $0.701$ & $0.818$ & $0.627$ & $1.000$ & $0.702$ & $0.815$ & $0.641$ \\
    \cmidrule(l){1-13}
    \rowcolor{priorrow} PD~\cite{DBLP:journals/corr/abs-2512-13317} & $0.946$ & $0.622$ & $0.062$ & $0.041$ & $0.906$ & $0.598$ & $0.049$ & $0.047$ & $0.890$ & $0.620$ & $0.136$ & $0.043$ \\
    \rowcolor{priorrow} HD~\cite{DBLP:journals/corr/abs-2512-13317} & $0.933$ & $0.649$ & $0.049$ & $0.019$ & $0.906$ & $0.599$ & $0.007$ & $\mathbf{0.015}$ & $0.875$ & $0.597$ & $\mathbf{0.006}$ & $\mathbf{0.007}$ \\
    \cmidrule(l){1-13}
    \rowcolor{ourrow} CP & $0.000$ & $0.000$ & $0.017$ & $\mathbf{0.002}$ & $0.652$ & $0.488$ & $0.122$ & $0.082$ & $0.475$ & $0.440$ & $0.241$ & $0.169$ \\
    \rowcolor{ourrow} EC & $\mathbf{0.951}$ & $\mathbf{0.683}$ & $0.054$ & $0.073$ & $\mathbf{0.985}$ & $0.654$ & $0.171$ & $0.038$ & $\mathbf{0.985}$ & $0.657$ & $0.168$ & $0.038$ \\
    \rowcolor{ourrow} OF & $0.694$ & $0.553$ & $\mathbf{0.000}$ & $0.020$ & $0.933$ & $\mathbf{0.671}$ & $\mathbf{0.003}$ & $0.038$ & $0.982$ & $\mathbf{0.699}$ & $0.019$ & $0.037$ \\
    \bottomrule
  \end{tabular}
\end{table*}

The forget set of $S_2$ holds five times as many identities as the forget set of $S_1$, and this affects the confusers and the manifold hiders differently.
The confusers give the same forgetting at both scales, and what changes is the cost in performance on retain sets.
At $\mathrm{FPIR}=10^{-4}$ the retain TPIR decreases from $0.946$ to $0.185$ for the pairwise dispersion, whereas the retain TPIR for the hard dispersion decreases only from $0.933$ to $0.912$ (\Cref{tab:results}).
At $S_2$ the cross-forget FMR between two forget-train images decreases to $0.000$ for the centroid push, against $0.265$ at $S_1$, so distinct forget identities stay linkable only at the smaller scale.

The manifold hiders assign one target to each forget image, so the larger forget set requires five times as many targets. Extended CosFace and the orthonormal frame differ in whether the targets are still fitted.
Extended CosFace, whose targets are trainable rows of the CosFace head, places fewer of the forget-train embeddings at their assigned targets, and the forget-train TMR increases from $0.001$ to $0.153$ at $\mathrm{FMR}=10^{-4}$.
The targets of the orthonormal frame are fixed before the optimisation, and the forget-train TMR for the orthonormal frame stays at $0.009$.
The orthonormal frame pays for the larger forget set on the retained identities, whose TMR decreases from $0.985$ to $0.912$ and whose TPIR decreases from $0.694$ to $0.447$, at a footprint that grows from $0.10$ to $0.24$.

We also run the protocols on two Vision Transformer backbones~\cite{DBLP:conf/iclr/DosovitskiyB0WZ21}, ViT-S and ViT-B, trained on the same identity subset under the same CosFace head, and we fine-tune every forgetting method from them under the configurations of \Cref{ssec:training}.
Adversarial forgetting can therefore happen on other backbones. Orthonormal frame gives a forget-train TMR of $0.000$ on all three backbones, so the fixed targets are as effective in a transformer embedding space as in a convolutional one.
The forgetting of the hard dispersion also strengthens with the backbone size, and the TMR between two forget-eval images decreases from $0.128$ to $0.055$.

The type of backbone influences which group incurs the cost of the forgetting (Finding~\ref{fnd:backbone}).
The retain TPIR for the orthonormal frame increases from $0.694$ on the iResNet-34 to $0.982$ on ViT-B, whereas the retain TPIR for the hard dispersion decreases from $0.933$ to $0.875$, so the ordering of the confusers and the manifold hiders on the retained identities is reversed (\Cref{tab:backbones}).
The manifold hiders pay on the forget comparisons, where the TMR between a forget-train and a forget-eval image increases from $0.005$ to $0.022$ for the orthonormal frame and from $0.044$ to $0.110$ for extended CosFace.

\subsection{Cost, Trade-offs, and Recommendations}
\label{ssec:results-cost}

Our findings suggest that there is no single Pareto-optimal forgetting method among the methods of this paper, and the choice of the adversarial forgetting method depends on the use case (Finding~\ref{fnd:verdict}).
The confusers and the manifold hiders both give weak forgetting at a small cost in verification, whereas only the orthonormal frame gives strong forgetting, which is paid for in the shape of the retain score distributions.
Neither kind of forgetting is measured by a single cost, since the centroid push loses no test TMR and still leaves the retain identities unidentifiable at $\mathrm{FPIR}=10^{-4}$, and since the footprint counts a tightening and a broadening of the distribution of mated retain scores equally (\Cref{ssec:results-footprint}).

A method is dominated when another gives more forgetting at no greater cost, and the methods that are not dominated form the Pareto front of their panel.
\Cref{fig:pareto} reports the $S_1$ forget set.

Four of the five methods are on the weak-forgetting front at $S_1$, and the greatest loss in test TMR anywhere on that front is $0.033$, so weak forgetting barely constrains the choice.
The orthonormal frame is dominated at $S_2$, where the hard dispersion reduces the forget-eval TMR further at a loss of $0.013$ against $0.027$, and the pairwise dispersion is dominated at both scales.
The choice is made in the strong-forgetting panels, since the orthonormal frame is the only method that brings the mated forget-train scores onto the non-mated-test distribution, at a $W_1$ distance of $0.006$ against the $0.100$ of the hard dispersion.
The orthonormal frame pays a footprint of $0.10$ against $0.015$ for that distance, and the same ordering holds at $S_2$.
The centroid push and the dispersion losses train without a classifier head at the lowest cost, whereas the two per-image-target methods require four times the parallelism, extended CosFace at the GPU-hours of the dispersion losses and the orthonormal frame at about $1.6$ times as many.
The per-image-target methods store one target per forget image, so their memory grows with the forget set.

Where the forgetting must extend to images on which the losses never operated, and the comparisons of a forgotten identity must become indistinguishable from the comparisons between unrelated identities, the orthonormal frame is the method of choice.
The cost in retain identification for the orthonormal frame is confined to the iResNet-34 (\Cref{ssec:results-scale}), and the footprint grows with the forget scale, from $0.10$ at $S_1$ to $0.24$ at $S_2$.
Where unlinkability at the operating point is enough, the hard dispersion is the method of choice, on the Pareto front of both panels at both scales and at a footprint on $S_1$ of $0.015$ against the $0.10$ of the orthonormal frame.
The remaining methods each fail at least one requirement, extended CosFace because an averaged forget-train enrolment re-links the forget-eval probes, the centroid push because distinct forget identities stay linkable at $S_1$ and the retain identification is costly, and the generic baselines because NegGrad+ fails the forgetting and CURE the retention (\Cref{ssec:results-generic}).

\section{Conclusions and Future Work}
\label{sec:conclusions}

We have framed adversarial forgetting as making chosen identities unlinkable across their capture instances while the rest of the population stays linkable. We distinguished weak forgetting, which pushes the mated forget comparisons below the threshold, from strong forgetting, which brings them onto the distribution of comparisons between unrelated identities. We proposed three forgetting losses and evaluated them against the pairwise and hard dispersion losses, NegGrad+, and CURE, in 1:1 verification and 1:N identification, at two forget scales and on three backbones.

The effects of these losses fall into two categories: confusers (the pairwise dispersion, the hard dispersion, and the centroid push) and manifold hiders (extended CosFace and the orthonormal frame). The confusers give weak forgetting at a cost of at most $0.033$ in test TMR, and two images of different forgotten identities are still matched to one another more often than two images of unrelated identities. The manifold hiders move the mated forget comparisons onto the non-mated scores, and only the orthonormal frame produces strong forgetting, which holds where a forget-train image is involved in the comparison, that is, where at least one image was seen by the altered model. We also considered an oracle that supplies the linkage the altered model no longer provides, through which an averaged enrolment template brings part of the identity information back for every method, and concluded that the forgetting stays incomplete while a gallery still holds templates built from the linked images. Adversarial forgetting also leaves a footprint in the retain score distributions, and an adversary who knows how the comparison scores of the face recognition model are distributed can measure that footprint without any knowledge of the forget set, so a large footprint indicates that the model in service has been altered. Of the generic baselines, NegGrad+ leaves the forget-train TMR at $0.998$ against the $0.912$ of the face recognition model, so the forget identities stay as verifiable after the forgetting as before. The training under CURE produces degenerate models that no longer identify the retain identities.

Our results point to the orthonormal frame as the method of choice when the forgetting must extend to the images on which the losses never operated, at a cost in retain identification that appears on the iResNet-34 and not on the transformers, and at a footprint that grows with the forget scale. The hard dispersion is the better choice when unlinkability at the operating point is enough, at a footprint of $0.015$ on $S_1$ against the $0.10$ of the orthonormal frame.

The cost in retain identification for the orthonormal frame decreases from $0.208$ on the iResNet-34 to $0.018$ on ViT-B at $\mathrm{FPIR}=10^{-4}$, so the costs we report may not be fixed properties of the losses. Our results are nonetheless for one margin loss on a 100,000-identity subset of WebFace4M, and whether these mechanisms carry to other margin losses and other corpora is an open question. We report two forget scales, at 1\% and 5\% of the training population, and whether the forget rates stay low as the forget set grows towards half the population is a question for scales beyond $S_2$.

\section*{Acknowledgments}

This work has received funding from the European Union's Horizon Europe research and innovation programme under Grant Agreement No.~101189650 (CERTAIN: Certification for Ethical and Regulatory Transparency in Artificial Intelligence), and the Swiss State Secretariat for Education, Research and Innovation (SERI).

\bibliographystyle{IEEEtran}
\bibliography{refs}

\end{document}